\documentclass[runningheads]{llncs}
\usepackage[T1]{fontenc}
\usepackage{graphicx}
\usepackage{booktabs}
\usepackage[misc]{ifsym}
\newcommand{\corr}{(\Letter)}
\usepackage{mwe}

\usepackage{hyperref} 
\usepackage{ctable}
\usepackage{pifont}
\usepackage[table,usenames,dvipsnames]{xcolor}
\definecolor{lightgray}{gray}{0.95}
\usepackage{amsmath}
\usepackage{amssymb}
\usepackage{multicol}
\usepackage{multirow}
\usepackage{graphicx}
\usepackage{comment}
\usepackage{arydshln}
\usepackage[table]{xcolor}
\usepackage{subcaption} 
\usepackage[outercaption]{sidecap}
\usepackage{xurl}

\begin{document}
%
\title{Hierarchical Spatio-Temporal Transformer for Coherent Emergency Department Forecasting} 
\titlerunning{Hierarchical Spatio-Temporal Transformer for Coherent ED Forecasting}
%
\author{Filipa Lino\inst{1} \corr \and Bárbara Tavares\inst{1} \and
Carlos Santiago\inst{1} \and Cláudia Soares\inst{2} \and
Manuel Marques\inst{1}}

\authorrunning{F. Lino et al.}
%

\institute{Institute for Systems and Robotics, LARSyS, Instituto Superior Técnico, Portugal  \and
NOVA School of Science and Technology, Caparica, Portugal\\
\email{filipa.lino@tecnico.ulisboa.pt}}

\maketitle              
\begin{abstract}

Emergency Departments (EDs) are critical access points in healthcare systems, yet they face persistent pressure from unpredictable patient demand, seasonal surges, and non-urgent visits. Effective ED planning requires forecasts at multiple decision-making levels: hospitals need local demand estimates for staffing and bed management, regions require forecasts to coordinate healthcare units, and national authorities need system-wide projections for capacity planning. However, most existing approaches forecast ED demand independently at a single level, ignoring the hierarchy linking hospitals, regions, and national systems. This can produce incoherent predictions, where hospital-level forecasts do not aggregate consistently to regional or national demand. We propose HierSTT, a hierarchical Transformer-based framework for coherent multi-level ED forecasting. HierSTT jointly predicts hospital, regional, and national level demand in a single end-to-end model. A Temporal Fusion Transformer captures national dynamics, while spatio-temporal Transformer encoder-decoder modules model regional and hospital demand conditioned on higher-level forecasts. A coherence-aware loss penalizes cross-level inconsistencies during training. We further introduce a nationwide Portuguese ED dataset covering 81 hospitals across 5 regional health administrations, with heterogeneous covariates at each level. Experiments show that HierSTT reduces average WAPE by 32\% relative to the best non-hierarchical deep learning baseline and outperforms all classical hierarchical reconciliation methods, while producing near-coherent predictions across levels. Additional resources associated with this work are available at \url{https://github.com/FilipaLino/HierSTT}.



\keywords{Emergency Department Forecasting \and Hierarchical Time Series \and Transformers \and Spatio-Temporal Modeling \and Forecast Coherence.}
\end{abstract}
\section{Introduction}
\label{sec:introduction}

\begin{figure}[t]
    \centering
    \begin{subfigure}{0.49\textwidth}
        \centering
        \includegraphics[width=0.9\textwidth,trim={2mm 0 2mm 0}, clip]{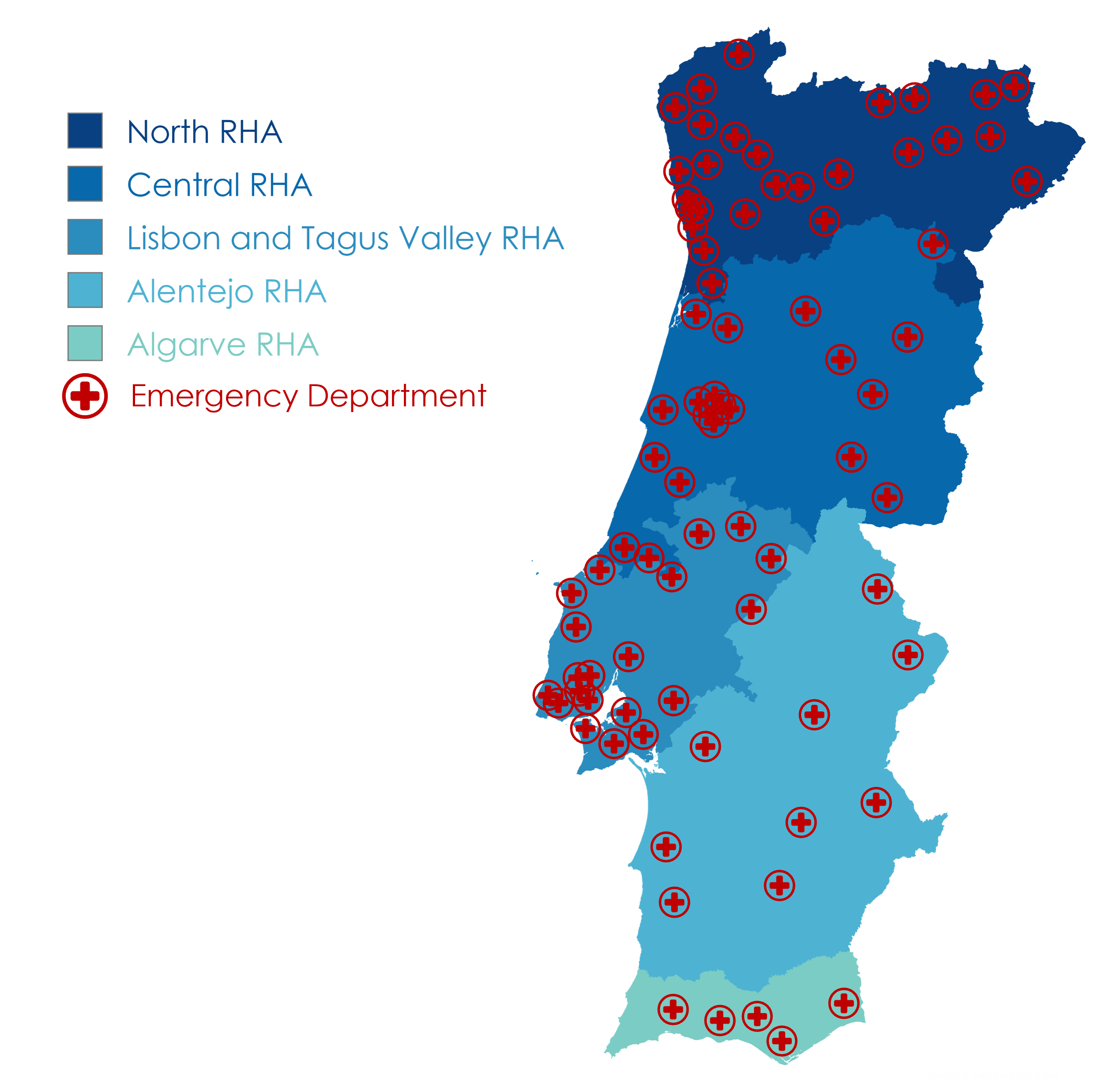}
        \caption{}
        \label{fig:map_portugal}
    \end{subfigure}
    \hfill
    \begin{subfigure}{0.49\textwidth}
        \centering
        \includegraphics[width=0.9\textwidth,trim={2mm 0 0 0}, clip]{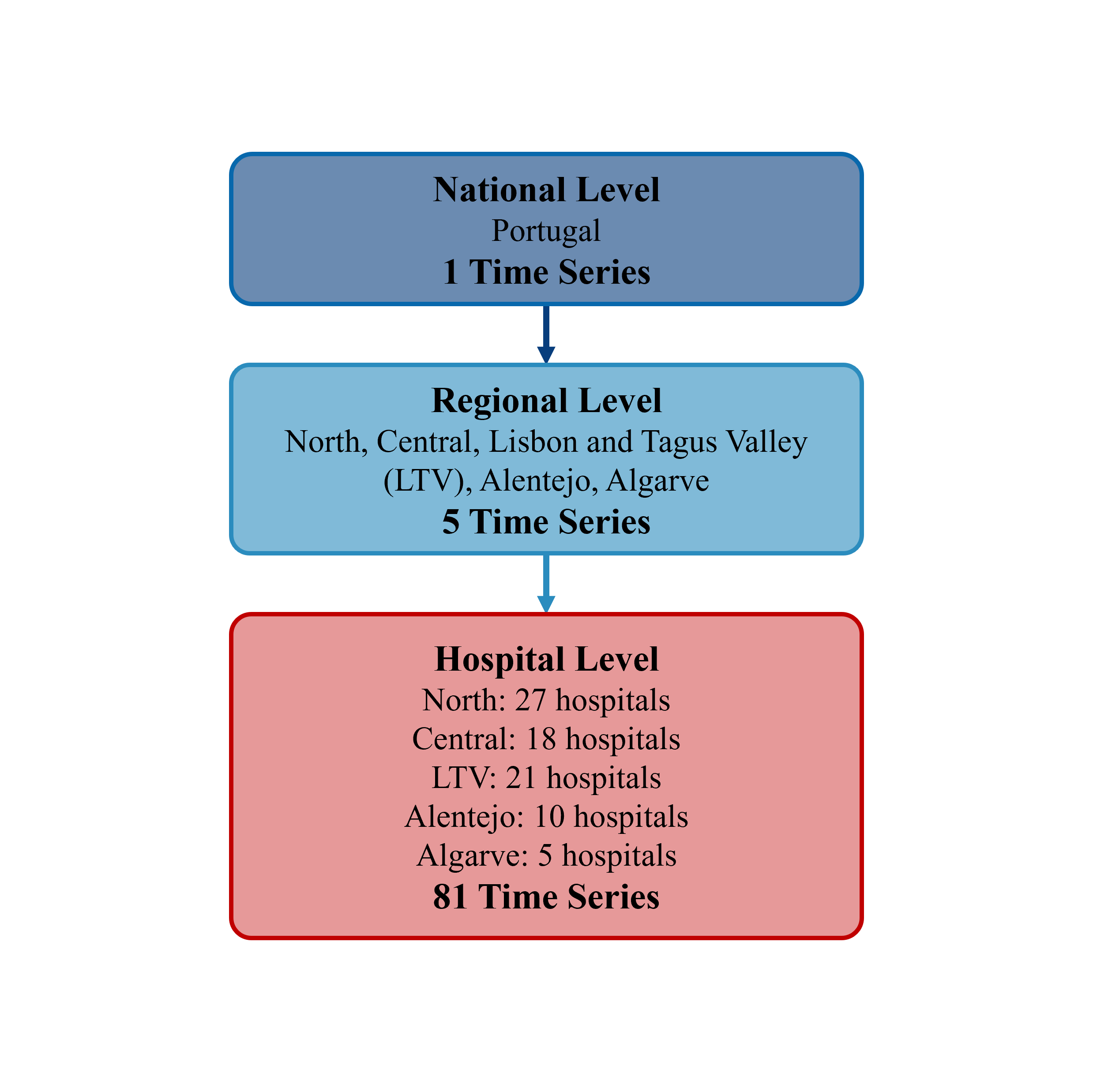}
        \caption{}
        \label{fig:hierarchy_distribution}
    \end{subfigure}
   \caption{Geographical and hierarchical organization of the ED forecasting dataset. (a) Spatial distribution of EDs in Portugal, grouped by RHAs. Each point represents a hospital. (b) Hierarchical structure of the dataset.}
   \label{fig:portugal_hierarchy}    
\end{figure}

Emergency Departments (EDs) play a critical role in the healthcare system, serving as the largest source of hospital admissions. These departments operate under persistent pressure due to unpredictable fluctuations in patient demand~\cite{turner2020effects}, which when exceeding operational capacity leads to overcrowding, a significant challenge~\cite{10.4103/JETS.JETS_42_19} causing negative patient outcomes, longer waiting times, increased costs, and staff burnout. Accurate forecasting of ED demand is therefore crucial for improving patient care, optimizing resource allocation, and supporting policy decision-making~\cite{savioli2022emergency}.

Forecasting ED visits is challenging due to seasonality, weekday effects, holidays, epidemics, weather patterns, and regional mobility dynamics~\cite{silva2023predicting}. Traditional approaches such as autoregressive integrated moving average (ARIMA), exponential smoothing (ETS), and other statistical methods have been widely applied~\cite{silva2023predicting,zhang2018base}, while more recently deep learning models, including long short-term memory (LSTM), and transformer-based architectures, have demonstrated improved capacity to capture nonlinear temporal dependencies~\cite{ming2025deep,145815,ming2026transformers}. Despite these advances, most existing studies formulate ED forecasting as an isolated single-level problem, focusing independently on one hospital or one region. 

However, healthcare systems naturally exhibit a hierarchical structure: individual hospitals belong to regional administrative units, which in turn compose the national healthcare network (Fig.~\ref{fig:portugal_hierarchy}). Ignoring this structure may lead to incoherent forecasts, where prediction generated independently at different levels are mutually inconsistent. Furthermore, data across levels is often heterogeneous. While hospital data may capture operational and demographic factors, regional and national levels provide broader environmental and population-level signals, making joint modeling desirable.

Hierarchical time-series forecasting addresses this challenge by producing consistent forecasts across aggregation levels~\cite{mancuso2021machine,10.1145/3616855.3635806}. Classical reconciliation methods adjust independently generated predictions post-hoc~\cite{olivares2023hierarchicalforecast}, but fail to exploit cross-level dependencies during learning. 


To address these limitations, we propose HierSTT, a hierarchical Transformer-based deep learning framework for coherent multi-level ED demand forecasting. The model simultaneously predicts patient visits over a 4-week horizon at hospital (81), Regional Health Areas (RHAs) (5 regions), and national levels of the Portuguese National Health Service. It combines a Temporal Fusion Transformer~\cite{caldas2022temporal} at the national level with spatio-temporal Transformer encoders and decoders~\cite{ji2024spatio} at regional and hospital levels. Predictions are generated in a top-down manner, enabling higher-level forecasts to inform lower-level estimates, simultaneously integrating historical demand with heterogeneous level-specific covariates. A coherence-aware loss penalizes inconsistencies between aggregated lower-level predictions and upper-level forecasts during training. Alongside the model, we introduce a nationwide Portuguese ED dataset reflecting real-world heterogeneous data availability across administrative levels.


Extensive experiments demonstrate that HierSTT achieves superior balance between predictive accuracy and hierarchical coherence compared to statistical baselines, non-hierarchical deep learning models, and reconciliation methods. In summary, our main contributions are the following:
\begin{itemize}
    \item A hierarchical Transformer-based framework for simultaneous ED demand forecasting across multiple levels;
    \item A coherence-aware training objective enforcing cross-level consistency;
    \item A new nationwide Portuguese ED dataset with heterogeneous, level-specific covariates. 
\end{itemize}


\section{Related Work}
\label{sec:related_work}
\subsection{Classical Approaches for ED Forecasting}

Classical time-series forecasting methods have long been used for demand prediction. Na\"ive forecasting assumes that future values follow the most recent observations, providing a strong baseline in persistent series~\cite{hyndman2018forecasting}. Yule et al.~\cite{yule1971method} introduced auto-regressive (AR) and moving-average (MA) processes, later unified under the ARIMA framework~\cite{box2015time}, which models temporal dependencies through linear relationships in past observations and residuals. Exponential Smoothing methods, such as ETS by Hyndman et al.~\cite{hyndman2002state}, decompose time series into Error, Trend and Seasonality components, automatically selecting most suitable additive formulation for different temporal patterns~\cite{findley1998new}.

In ED patient volume forecasting, these methods have been adopted due to its strong temporal structure, which follows calendar-driven patterns such as weekdays, holidays, and seasonal trends~\cite{batal2001predicting}. Classical statistical approaches, such as moving averages and ARIMA, have therefore been used to model these dynamics~\cite{https://doi.org/10.1002/sim.4780071007}. Their interpretability and computational efficiency make them attractive for operational settings.

However, ED demand is influenced by complex and interactive factors, including patient demographics, epidemiological trends, environmental conditions, and operational characteristics. The nonlinear and dynamic nature of these relationships challenges the assumptions of linearity and stationarity of classical models, often limiting their predictive performance in real-world scenarios~\cite{KAYACAN20101784}. 


\subsection{Deep Learning for Time Series}
Deep learning (DL) has become a major research direction in time series forecasting due to its ability to learn complex nonlinear patterns and multivariate interactions~\cite{10.1080/01605682.2022.2118629}. In ED forecasting, Kadri et al.~\cite{kadri2020rnn} applied Recurrent Neural Networks (RNNs) to patient-flow prediction, but RNNs often suffer from training instability and limited ability to capture long-range temporal dependencies.

To address these limitations, more advanced recurrent architectures have been proposed.  Kashani et al.~\cite{145815} showed that Long Short-Term Memory (LSTM) networks improve performance in highly correlated temporal settings. Building on recurrent models, sequence-to-sequence (Seq2seq) architectures became relevant for forecasting by encoding historical information into latent representations and decoding future horizons. Seq2Seq models based on Gated Recurrent Units (GRUs) and LSTMs have shown strong performance in capturing long-range temporal dependencies and multistep forecasting patterns~\cite{xu2021fm,masood2022multi}.


A key advantage of DL models is their ability to integrate multiple exogenous variables. Several studies have shown that calendar-related features, weather conditions, and air quality influence ED demand variability~\cite{caldas2022temporal,Boyle358,alvarez2024evaluating}. These findings have motivated the use of attention mechanisms to dynamically weight relevant inputs. Chaves et al.~\cite{alvarez2024evaluating} demonstrated the effectiveness of attention-based architectures for forecasting ED patient admissions.

Transformer-based architectures have advanced time series forecasting through self-attention. In the ED domain, Ming et al.~\cite{ming2026transformers} demonstrated the strong predictive performance of Transformers, while Ji et al.~\cite{ji2024spatio} proposed a spatio-temporal variant for jointly modeling temporal and spatial dependencies. Temporal Fusion Transformers (TFT) have shown promising results in healthcare, with Pulkkinen et al.~\cite{pulkkinen2020forecasting} applying TFT to short-term hospital demand prediction and Caldas et al.~\cite{caldas2022temporal} extending it to longer-term forecasting across Portugal's RHAs.

Alongside Transformer-based approaches, alternative deep forecasting architectures have also emerged. Oreshkin et al.~\cite{oreshkin2019n} introduced N-BEATS, a deep residual architecture that achieved strong forecasting performance without recurrent or attention mechanisms. However, despite these advances, most DL-based approaches remain designed for single-level forecasting and do not explicitly model hierarchical dependencies across healthcare system levels.

\subsection{Hierarchical Time Series Forecasting}
Hierarchical Time Series (HTS) forecasting generates predictions across multiple aggregation levels while ensuring coherence, \textit{i.e.}, that lower-level forecasts aggregate consistently to higher-level ones. HTS has been applied in domains such as tourism demand forecasting across national and regional levels~\cite{olivares2023hierarchicalforecast} and retail sales forecasting across items, departments, and product categories~\cite{Nasios_2022}.

Classical HTS approaches include bottom-up, top-down, middle-out,
and reconciliation-based strategies, which usually generate forecasts independently before enforcing coherence through post-hoc aggregation constraints~\cite{olivares2023hierarchicalforecast}. However, post-hoc reconciliation limits the ability to learn cross-level dependencies and may propagate errors across the hierarchy.

Recent work have sought to integrate reconciliation into model training. Mancuso et al.~\cite{mancuso2021machine} proposed a deep neural network with a reconciliation layer and a coherence-aware loss, reconciling lower-level predictions using outputs from immediately higher level, through sequential training across hierarchical level pairs. Wang et al.~\cite{10.1145/3616855.3635806} introduced the Neural Reconciler, combining encoder-decoder architectures with Normalizing Flows to model hierarchical forecasts. However, this method still follows a multi-stage pipeline, where base forecasts are first generated and later refined through reconciliation. Thus, hierarchical constraints are not learned jointly with temporal representations, potentially leading to suboptimal performance. 

Overall, while ED forecasting has advanced through statistical and deep learning approaches, the integration of hierarchical structure and heterogeneous covariates remains underexplored. To the best of our knowledge, no prior work jointly addresses hierarchical coherence and heterogeneous multi-level covariates in ED forecasting.

\section{Dataset}
\label{sec:dataset}
\subsection{Data Sources and Collection}

This work is based on a nationwide dataset of ED activity in Portugal, comprising historical records collected across multiple administrative levels of the healthcare system. Data were initially available for 91 hospitals. Ten recently established hospitals were excluded due to insufficient data, resulting in a final dataset of 81 hospitals, grouped into 5 RHAs, as well as aggregated national-level indicators. This hierarchical structure is detailed in Fig.~\ref{fig:hierarchy_distribution}. The data span the period from January 1, 2021 to April 20, 2024. 

The Portuguese Ministry of Health provides open-source data on 
the National Health Service (SNS) activities in the \textit{Transparência} Website~\footnote{\url{https://transparencia.sns.gov.pt/explore/?sort=modified}} and daily ED monitoring platform~\footnote{\url{https://www.sns.gov.pt/monitorizacao-do-sns/servicos-de-urgencia/}}. These sources were combined and cross-referenced with external data, including Portugal air quality data~\footnote{\url{https://aqicn.org/map/portugal/pt/}}, Temperature~\footnote{\url{https://mesonet.agron.iastate.edu/request/download.phtml?network=PT\__ASOS}} and Mortality~\footnote{\url{https://evm.min-saude.pt/\#shiny-tab-a\_total}} to construct a unified dataset.


\subsection{Hierarchical Structure and Variables}
\noindent\textbf{Hospital Level}
At the hospital level, each record includes information such as the hospital unit (D0), locality, the total number of urgent episodes (M1), and detailed triage information according to the Manchester Triage System (M2 - blue, M3  - white, M4 - green, M5 - without triage, M6 - yellow, M7 - orange, M8 - red). Additional variables include average waiting time between triage and medical evaluation, type of emergency service (Basic Emergency Service - SUB, Medical-Surgical Emergency Service - SUMC, Polyvalent Emergency Service - SUP and Trauma Center - SUPCT), patient access pathways (self-admission, SNS24, through primary health care warrant, over emergency doctors or hospital doctors referrals), and admitted age ranges. Calendar-related variables, such as day, weekday, month, and holiday indicators, are also included. Furthermore, an operational variable was introduced to indicate whether each ED was open on a given day, as some units occasionally close, resulting in zero recorded visits.

\noindent\textbf{Regional Level}
At the regional level, data include aggregated demand indicators, average waiting times, and environmental variables such as air quality index (AQI) and temperature. Due to limited availability of localized environmental measurements, AQI and temperature were aggregated at the RHA level. The AQI reflects concentrations of major pollutants, including ozone (O\(_3\)), particulate matter (PM2.5 and PM10), nitrogen dioxide (NO\(_2\)), and sulfur dioxide (SO\(_2\)), and was computed according to established guidelines~\cite{AQI}.

\noindent\textbf{National Level}
At the national level, the dataset includes system-wide indicators such as aggregated ED demand, average waiting times, and mortality rates, providing a macro-level view of the healthcare system pressure. 

The availability of covariates varies across hierarchical levels. Hospital-level data capture local operational and demographic characteristics, whereas regional and national levels provide complementary environmental and health indicators, such as temperature and mortality rates. This creates a heterogeneous feature space across the hierarchy, reflecting real-world data availability and motivating models capable of integrating level-specific information. To ensure dataset completeness over the study period, missing values at specific timestamps were filled using linear interpolation. 



\section{Methodology}
\label{sec:methodology}

This section presents our Hierarchical Spatio-Temporal Transformer (HierSTT) for multi-step ED demand forecasting across hospital, regional, and national levels. The task consists of jointly predicting patient visits over a 28-day horizon using the previous 42 days, while ensuring hierarchical coherence and effectively leveraging heterogeneous covariates.

HierSTT jointly models temporal dynamics and cross-level dependencies through a top-down architecture in which higher-level representations guide lower-level forecasts. A coherence-aware loss further enforces consistency across the hierarchy, enabling the model to capture both global demand trends and local variations.



\subsection{Problem Formulation}
We consider a hierarchical forecasting problem defined over a healthcare system organized into three aggregation levels: hospital, regional, and national. At the lowest level, let $\mathcal{H} = \{1, \dots, H\}$ denote the set of hospitals, grouped into a set of regions $\mathcal{R} = \{1, \dots, R\}$, which together form the national level.

For each hospital $h \in \mathcal{H}$, we observe a univariate time series of daily ED visits, denoted by $y_t^{(h)} \in \mathbb{R}_{0+}$ at time $t$. Aggregated time series at the regional and national levels are defined as the sum of their corresponding lower-level series. Specifically, for each region $r \in \mathcal{R}$, and at the national level,
\begin{equation}
    y_t^{(r)} = \sum_{h \in \mathcal{H}_r} y_t^{(h)}, \quad 
    y_t^{(n)} = \sum_{h \in \mathcal{H}} y_t^{(h)},\ \text{and} \quad  
    y_t^{(n)} = \sum_{r \in \mathcal{R}} y_t^{(r)}.
    \label{eq:hosp-reg-nat}
\end{equation}


In addition to the target series, we consider exogenous covariates that vary across hierarchy levels, with some variables only available at specific levels.
Given a fixed-length input sequence of $T_{\text{in}} = 42$ past observations, the goal is to predict the future values of the time series over a horizon of $T_{\text{out}} = 28$ days for all hierarchy levels. Formally, for each time $t$, the model receives as input:
\begin{equation}
\mathbf{X}_{t-T_{\text{in}}:t} = \{\mathbf{x}_\tau^{(h)}, \mathbf{x}_\tau^{(r)}, \mathbf{x}_\tau^{(n)}\}_{\tau = t-T_{\text{in}}}^{t},
\end{equation}
where $\mathbf{x}_\tau^{(\cdot)}$ represents the observed features (including past visits) at each level.

The objective is to jointly predict hospital, regional, and national level demand over the forecasting horizon, while ensuring hierarchical coherence, \textit{i.e.}, lower-level forecasts must aggregate consistently to higher-level predictions according to Eq.~\eqref{eq:hosp-reg-nat}. To address this problem, we formulate hierarchical forecasting as a joint mapping from multi-level historical observations to future predictions across all levels:
\begin{equation}
\left( \hat{\mathbf{y}}_{t+1:t+T_{\text{out}}}^{(n)},\  \hat{\mathbf{y}}_{t+1:t+T_{\text{out}}}^{(r)},\  \hat{\mathbf{y}}_{t+1:t+T_{\text{out}}}^{(h)} \right)
= f_{\theta}\left( \mathbf{X}_{t-T_{\text{in}}:t} \right),
\end{equation}
for all hospitals $h \in \mathcal{H}$, regions $r \in \mathcal{R}$, and the national level, where $f_{\theta}$ models temporal dependencies and cross-level interactions.


To explicitly model hierarchical dependencies, forecasting is performed in a top-down manner, where national forecasts are first generated, followed by regional forecasts conditioned on national predictions, and hospital forecasts conditioned on regional predictions:
\begin{align}
\hat{\mathbf{y}}_{t+1:t+T_{\text{out}}}^{(n)}  &= 
f_{\theta}^{(n)}\left( \mathbf{X}_{t-T_{\text{in}}:t}^{(n)} \right), \label{eq:tft} \\
\hat{\mathbf{y}}_{t+1:t+T_{\text{out}}}^{(r)} &= 
f_{\theta}^{(r)}\left( \mathbf{X}_{t-T_{\text{in}}:t}^{(r)}, 
\hat{\mathbf{y}}_{t+1:t+T_{\text{out}}}^{(n)} \right), \\
\hat{\mathbf{y}}_{t+1:t+T_{\text{out}}}^{(h)} &= 
f_{\theta}^{(h)}\left( \mathbf{X}_{t-T_{\text{in}}:t}^{(h)}, 
\hat{\mathbf{y}}_{t+1:t+T_{\text{out}}}^{(r)} \right).
\end{align}

The national component $f_{\theta}^{(n)}$ is implemented using a TFT~\cite{caldas2022temporal} to capture global temporal patterns and heterogeneous covariates. Regional and hospital components, $f_{\theta}^{(r)}$ and $f_{\theta}^{(h)}$, use spatio-temporal Transformer encoder--decoder architectures~\cite{ji2024spatio} to jointly model temporal dynamics and cross-entity interactions. Higher-level predictions are incorporated into the decoding process, enabling lower-level forecasts to be conditioned on national and regional signals through attention mechanisms.

\begin{figure}[t]
    \centering
    \includegraphics[width=1\linewidth]{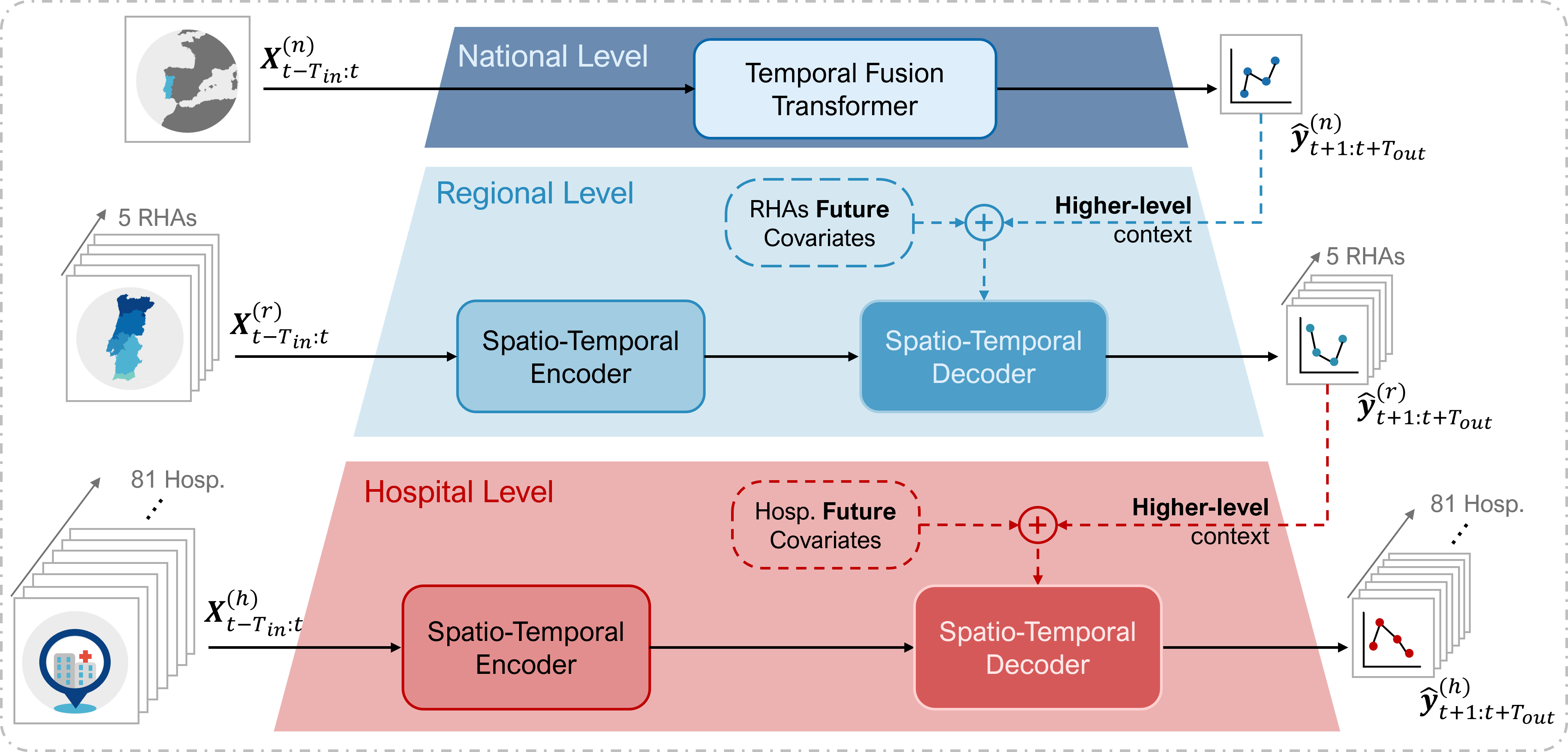}
    \caption{Overview of the HierSTT. A TFT models national-level ED demand, while spatio-temporal Transformer encoder--decoder modules forecast regional and hospital-level demand. Predictions are generated in a top-down manner: national forecasts condition regional decoding, and regional forecasts condition hospital decoding. }
    \label{fig:architecture}
\end{figure}

This formulation allows the model to jointly integrate heterogeneous covariates and hierarchical dependencies through top-down conditioning. A schematic overview of the proposed architecture is shown in Fig.~\ref{fig:architecture}.

\subsection{Temporal Fusion Transformer for National Forecasting}

At the national level, the model \(f_\theta^{(n)}\) is implemented as a TFT~\cite{caldas2022temporal}, which is designed for multi-horizon forecasting with heterogeneous covariates. The model processes a window of  \(T_{in}+T_{out}\) days, encoding the past \(T_{in}=42\) days through an LSTM and decoding the future \(T_{out}=28\) days using a second LSTM conditioned on static context, followed by multi-head self-attention over the decoded sequence. Input covariates include static variables, time-varying real-valued variables observed in the past (e.g., waiting times, ED demands, and mortality counts), and time-varying categorical variables that are known at future time steps (e.g., month, weekday, and holiday indicators).


Input variables are processed through Variable Selection Networks (VSNs), which learn time-dependent weights over covariates and suppress less relevant signals. Static covariates are encoded into context vectors that condition both variable selection and subsequent recurrent layers. However, since no natural static covariates exist at the national level, a dummy static feature is used for compatibility with the TFT architecture. The final output is a national forecast \(\hat{\mathbf{y}}^{(n)}_{t+1:t+T_{out}}\in \mathbb{R}^{T_{out}}\), which is then used to condition regional forecasting.




\subsection{Spatio-Temporal Transformer for Regional and Hospital Forecasting}

\begin{figure}[t]
    \centering
    \includegraphics[width=0.6\linewidth]{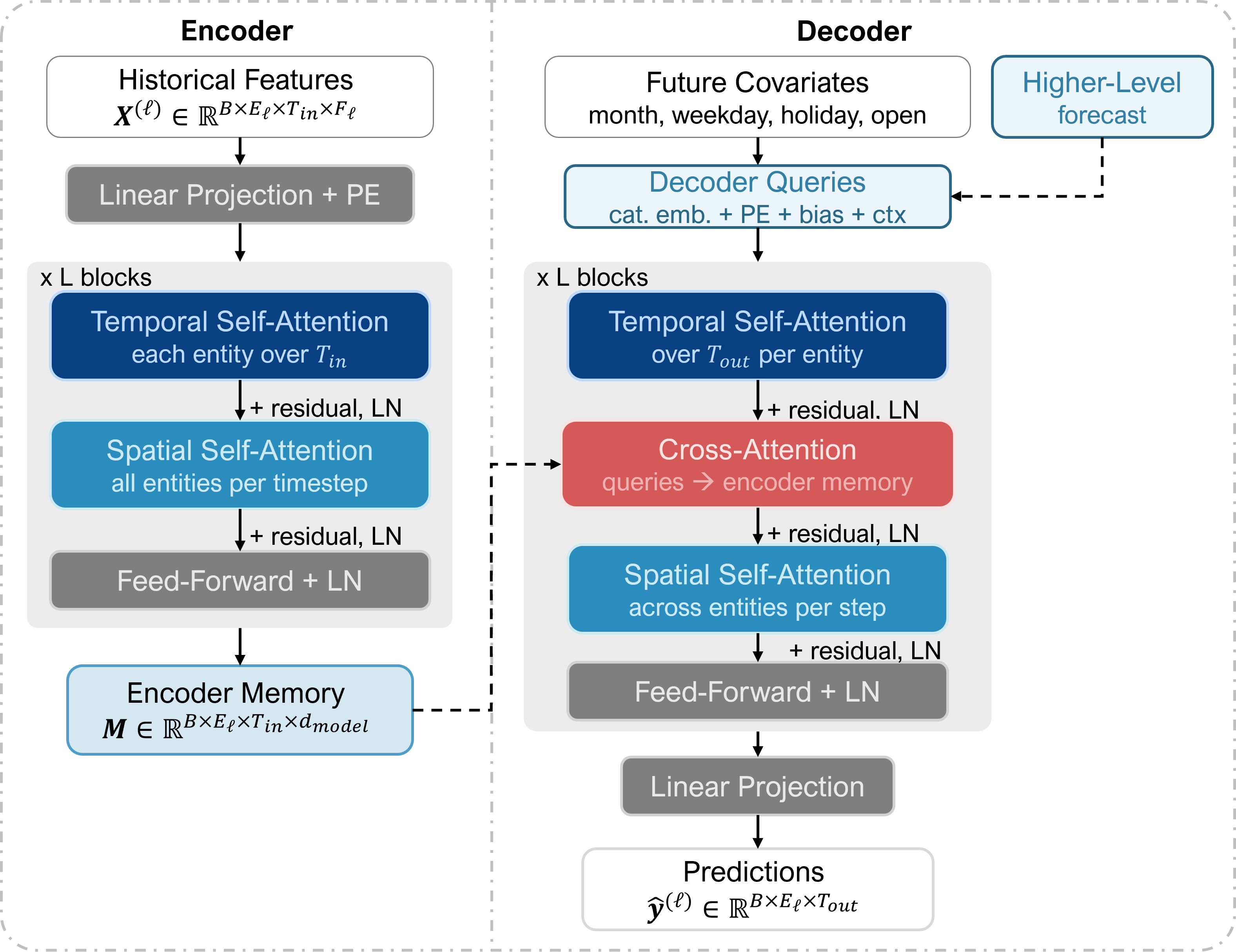}
    \caption{Spatio-temporal Transformer encoder--decoder used for regional and hospital forecasting. The encoder alternates temporal self-attention within entities and spatial self-attention across entities. The decoder incorporates future-known covariates and higher-level forecasts through temporal, cross-, and spatial-attention mechanisms to generate predictions \(\hat{\mathbf{y}}^{(\ell)}\).}
    \label{fig:st-transformer}
\end{figure}

At regional and hospital levels, the functions \(f_\theta^{(r)}\) and \(f_\theta^{(h)}\) are implemented as spatio-temporal Transformer encoder–decoder modules~\cite{ji2024spatio}. Unlike the national model, these levels include multiple interacting entities (5 regions and 81 hospitals). The architecture therefore models both temporal dependencies within each entity and spatial dependencies across entities, as shown in Fig.~\ref{fig:st-transformer}.

\subsubsection{Spatio-Temporal Encoder}

For each level \(\ell\in\{r,h\}\), the encoder receives historical features
\(\mathbf{X}^{(\ell)}\in\mathbb{R}^{B\times E_\ell \times T_{in} \times F_\ell}\), where \(B\) is the batch size, \(E_\ell\) the number of entities, and \(F_\ell\) the number of features. Inputs are projected to dimension \(d_{\text{model}}\), enriched with temporal positional encodings, and processed by \(L\) spatio-temporal blocks.

Each block alternates between two types of attention. Temporal self-attention is applied independently to each entity over the 42-day history. Spatial self-attention is then applied across all entities at each time step. This allows the encoder to learn both intra-entity temporal patterns and inter-entity correlations. Residual connections, layer normalization, and feed-forward layers are applied throughout.



\subsubsection{Spatio-Temporal Decoder}

The decoder predicts the next \(T_{out}=28\) days using the encoded memory, future-known covariates, and higher-level forecasts. Decoder queries are built from embeddings of month, weekday, holiday, and open status indicators, together with positional encodings. This yields a query tensor \(\mathbf{Q}\in\mathbb{R}^{B\times E_\ell \times T_{out} \times d_{model}}\). 


Higher-level predictions are projected and added to the decoder queries: national forecasts \(\hat{\textbf{y}}^{(n)}\in\mathbb{R}^{B \times T_{out}}\) condition regional decoding, while the corresponding regional forecasts \(\hat{\textbf{y}}^{(r_h)}\in\mathbb{R}^{B \times T_{out}}\), where \(r_h\) denotes the region which hospital \(h\) belongs, condition hospital decoding. 
Each decoder layer applies temporal self-attention over the prediction horizon within each entity, cross-attention to the encoder memory, and spatial self-attention across entities. A final linear projections yields the regional and hospital forecasts.

\subsection{Hierarchical Coherence-Aware Loss}

To encourage hierarchical consistency without hard-constraining the outputs, we add a soft coherence term to the training objective:


\begin{equation}
    \mathcal{L}= (1-\alpha)\left(\mathcal{L}^{(n)} + \mathcal{L}^{(r)} + \mathcal{L}^{(h)}\right) + \alpha \mathcal{L}_{coh},
    \label{eq:loss_general}
\end{equation}
where, \(\alpha \in [0,1]\) balances forecasting accuracy and hierarchical coherence, and each direct loss \(\mathcal{L}^{(.)}\) is computed using Smooth L1 loss. The coherence loss \(\mathcal{L}_{coh}\) penalizes inconsistencies between predictions across hierarchy levels. Specifically, we impose three aggregation constraints:
{\small
\begin{equation}
    \mathcal{L}_{coh} = \mathcal{L}\left(\hat{\mathbf{y}}^{(n)}, \sum_{r \in \mathcal{R}} \hat{\mathbf{y}}^{(r)}\right) + \mathcal{L}\left(\hat{\mathbf{y}}^{(n)}, \sum_{h \in \mathcal{H}} \hat{\mathbf{y}}^{(h)}\right) + \frac{1}{R}\sum_{r=1}^{R} \mathcal{L}\left(\hat{\mathbf{y}}^{(r)}, \sum_{h \in \mathcal{H}_r} \hat{\mathbf{y}}^{(h)}\right),
    \label{eq:loss}
\end{equation}
}\noindent for \(R=5\) regions. The first two terms enforce national-level consistency with aggregated regional and hospital forecasts, while the third enforces regional consistency with the corresponding hospital aggregates. Importantly, \( \mathcal{L}_{coh}\) compares predictions across levels rather than comparing aggregated predictions with ground truth, avoiding redundancy with the direct forecasting losses.




\section{Experimental Setup}
\label{sec:exp_setup}


The proposed framework is evaluated against statistical, deep learning, and hierarchical reconciliation baselines across hospital, regional, and national forecasting levels. Experiments assess both forecasting accuracy and hierarchical coherence using real-world ED demand data from the Portuguese National Health Service. Results demonstrate that the proposed model consistently outperforms competing approaches on both dimensions. All experiments were implemented in PyTorch and executed on an Intel Xeon Platinum 8260 CPU (2.40\,GHz) and an NVIDIA RTX A6000 GPU.

\subsubsection{Dataset and Splits}
Experiments were conducted on the Portuguese National Health Service ED dataset described in Section~\ref{sec:dataset}. Data prior to August 2021 were excluded 
to reduce the impact of COVID-19-related demand shifts. The remaining data were chronologically split into training, validation, and test sets using cutoff dates of 15 July 2023 and 2 December 2023, yielding 713, 140, and 140 days, respectively. No shuffling was applied to avoid temporal leakage.

All models used a 42-day encoder window (\(T_{in}\)) and a 28-day forecasting horizon (\(T_{out}\)). This configuration was chosen to capture seasonal trends and support effective resource management in hospital settings. Samples were generated using a sliding-window strategy with one day stride.



\subsubsection{Data Preprocessing}
All targets were log-transformed using \(\textbf{\~{y}}=\log (1+\textbf{y})\) to stabilize variance across hierarchical levels. Input features were level-specific, comprising 16, 17, and 26 features at the national, regional, and hospital levels, respectively. Features and targets were independently normalized using RobustScaler fitted on the training set. To compute coherence penalties \(\mathcal{L}_{coh}\) in the original count domain, predictions were inverse-scaled and mapped back via \(\hat{\mathbf{y}}=\mathrm{expm1}(\cdot)\) before evaluating aggregation constraints.

\subsection{Implementation Details}

The proposed model was instantiated with the following hyperparameters. At the national level, the TFT used a hidden dimension of 128, two LSTM layers, four attention heads, an embedding dimension of 8, and dropout rate of 0.1. The regional and hospital spatio-temporal Transformers used \(L=2\) blocks, \(d_{model}=128\), \(n_{head}=4\), and a dropout rate of 0.2.

The coherence weight was set to \(\alpha = 0.3\), which provided the best accuracy-coherence trade-off among tested values (detailed results are reported in the supplementary material~\ref{sec:alpha_ablation}). Models were trained for up to 200 epochs using AdamW with \(10^{-4}\) weight decay, OneCycleLR scheduling with peak learning rate of \(3\times10^{-4}\), gradient clipping at 1.0, and early stopping with 50-epoch patience based on validation loss. The best validation checkpoint was used for testing. For fair comparison, all deep learning experiments were repeated across six fixed random seeds, with results reported as mean \(\pm\) standard deviation.


\subsection{Baselines}
We compare the proposed model against statistical, deep learning, and hierarchical reconciliation baselines. Statistical baselines include Na\"ive~\cite{hyndman2018forecasting}, ARIMA~\cite{box2015time}, and ETS~\cite{hyndman2002state}, applied independently to each series. Non-hierarchical deep learning baselines include LSTM~\cite{ming2025deep}, Seq2Seq-LSTM~\cite{masood2022multi}, Seq2Seq-GRU~\cite{xu2021fm}, Transformer~\cite{vaswani2017attention}, TFT~\cite{caldas2022temporal}, and N-BEATS~\cite{oreshkin2019n}, all trained independently at each hierarchy level without coherence constraints. Finally, using the Nixtla framework~\cite{olivares2023hierarchicalforecast}, we evaluate Bottom-Up, Top-Down, and Middle-Out reconciliation strategies applied to the statistical forecasts. These baselines assess the benefit of jointly learning hierarchical dependencies.




\subsection{Evaluation Metrics}

Performance was evaluated in count space after inverting all preprocessing transformations. We report Mean Absolute Error (MAE), Root Mean Squared Error (RMSE), and Weighted Absolute Percentage Error (WAPE):
\begin{equation}
\text{WAPE} = \frac{\sum_t |\hat{y}_t - y_t|}{\sum_t y_t}\times 100.
\end{equation}

MAE and RMSE measure prediction error magnitude, with RMSE penalizing large errors more strongly. WAPE measures scale-normalized relative error. 

Hierarchical coherence was evaluated using the Hierarchical Aggregation Error (HAgE), computed as the WAPE between aggregated lower-level forecasts and higher-level predictions (Pred. HAgE). We also report HAgE against the observed higher-level ground truth (G.T. HAgE) to assess error propagation across hierarchy levels. Three aggregation settings were considered: hospital-to-regional, hospital-to-national, and regional-to-national. Perfectly coherent forecasts achieve Pred. HAgE~=~0.



\section{Results}
\label{sec:results}

\begin{figure}[t]
    \centering
    \includegraphics[width=\linewidth]{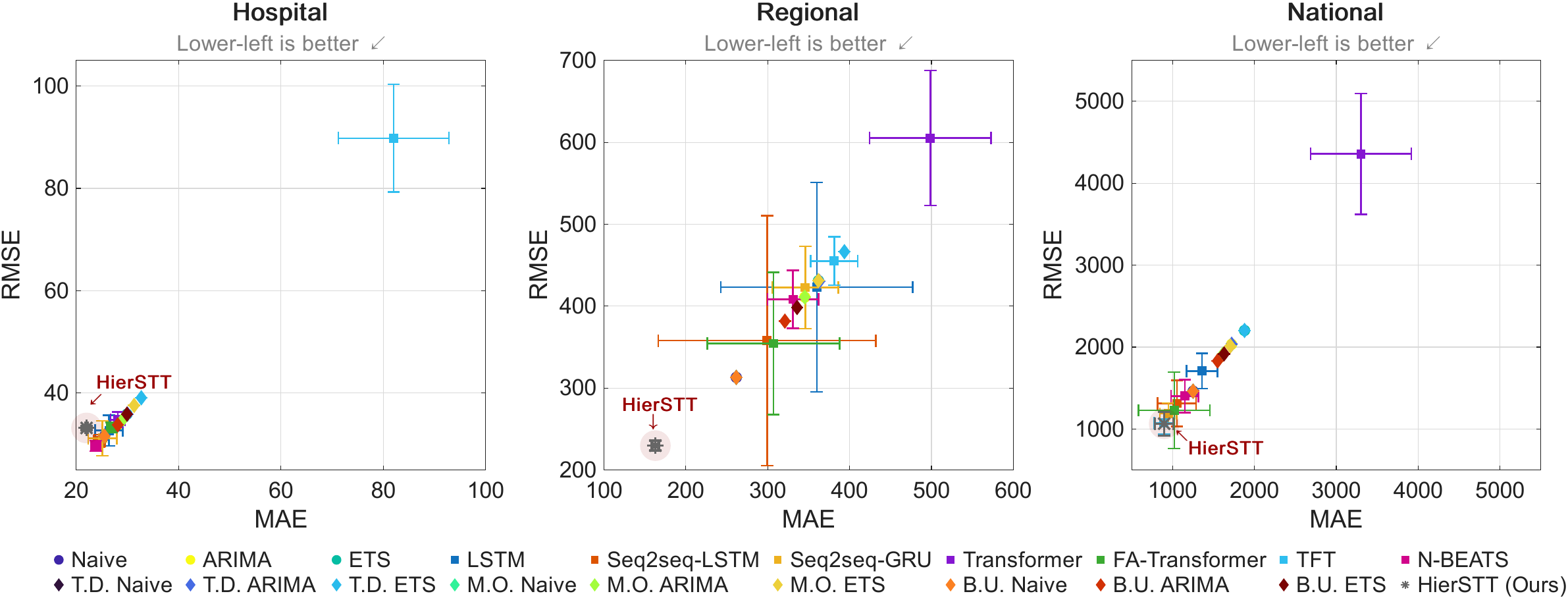}
    \caption{Mean MAE vs. mean RMSE for the hospital, regional, and national levels. Error bars denote standard deviation across seeds for deep learning models. T.D. = Top Down; M.O. = Middle Out; and B.U. = Bottom Up.}
    \label{fig:pred_accuracy}
\end{figure}

We evaluate HierSTT against all baselines across two complementary dimensions: forecasting accuracy and hierarchical coherence. Fig.~\ref{fig:pred_accuracy} reports absolute forecasting errors by level, while Fig.~\ref{fig:acc_vs_hage} summarizes the trade-off between accuracy and hierarchical coherence. Detailed numerical results per level are reported in the supplementary material~\ref{sec:results_detail}.

\subsection{Overall Forecasting Performance}

Fig.~\ref{fig:pred_accuracy} reports MAE and RMSE for hierarchy levels. At the hospital level, HierSTT achieves the lowest MAE, improving by approximately 8\% over N-BEATS, the second-best model. However, it has a slightly higher RMSE, indicating greater sensitivity to peak deviations in variable hospital demand. Although N-BEATS achieves the lowest hospital-level RMSE, this advantage does not extend to higher levels. At the regional level, HierSTT reduces MAE and RMSE by approximately 38\% and 27\%, respectively, over the second-best Na\"ive baseline. At the national level, its TFT top-level forecaster retains its standalone performance within the hierarchical architecture: HierSTT achieves the best RMSE and is essentially tied with TFT on MAE. Overall, HierSTT is the only model consistently positioned in the lower-left region across all three levels, while competing models tend to excel at one level at the expense of others.


\subsection{Hierarchical Coherence}

\begin{figure}[t]
    \centering
    \includegraphics[width=\linewidth]{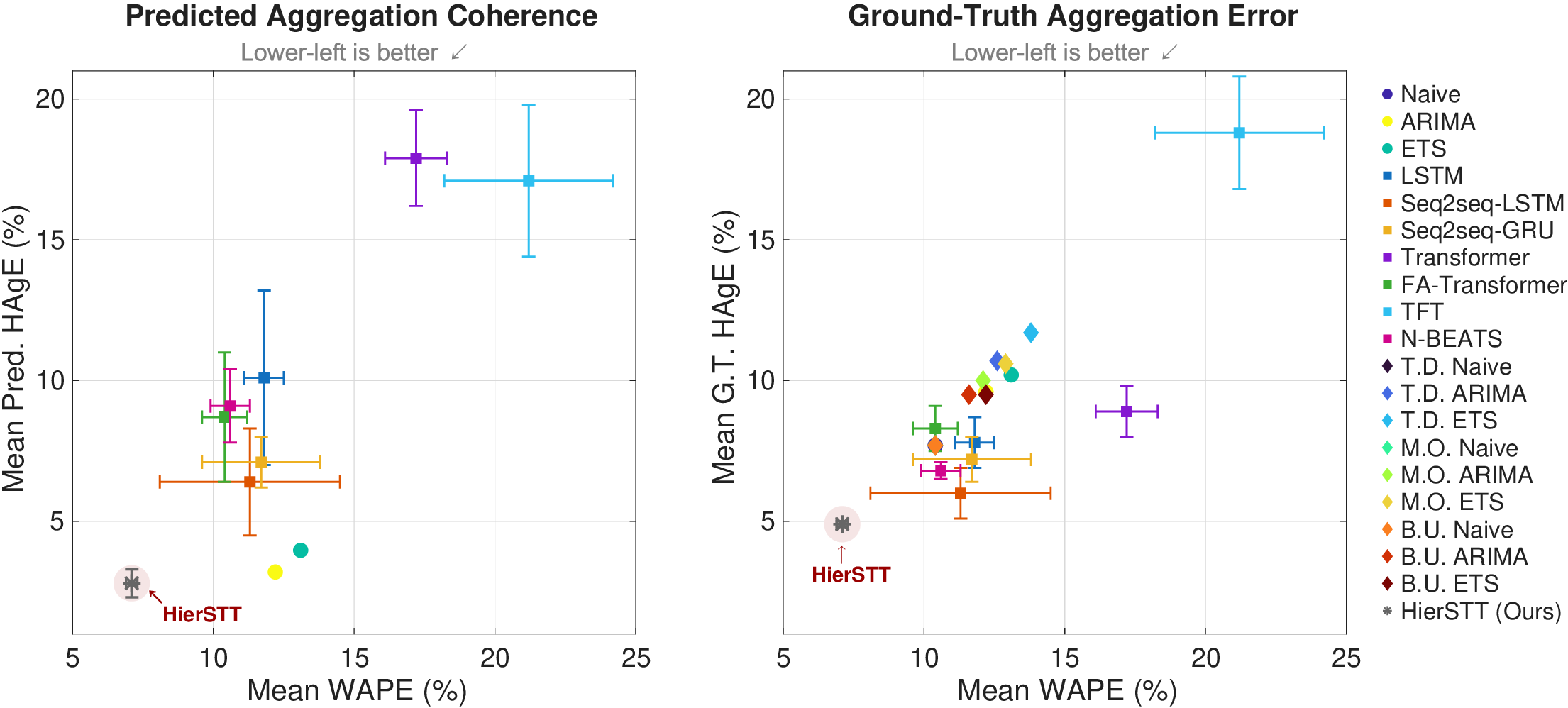}
    \caption{Trade-off between forecasting accuracy (mean WAPE) and hierarchical coherence (mean HAgE) across all models. The left plot measures coherence between aggregated lower-level and direct higher-level forecasts, while the right compares aggregated predictions against ground-truth aggregates. Results are averaged over hospital$\rightarrow$regional, hospital$\rightarrow$national, and regional$\rightarrow$national transitions.}
    \label{fig:acc_vs_hage}
\end{figure}

Coherence arises differently depending on the forecasting strategy. Na\"ive forecasts are implicitly coherent because they preserve the aggregation structure of historical observations, while post-hoc reconciliation methods enforce coherence through aggregation or disaggregation operations. Independently trained models ignore hierarchical consistency, whereas HierSTT learns coherence directly through its training objective.

Fig.~\ref{fig:acc_vs_hage} reports prediction-side HAgE against mean WAPE (left) and ground-truth HAgE against mean WAPE (right). Na\"ive and reconciliation-based models achieve zero prediction-side HAgE by construction and are therefore excluded from the left plot. Among unconstrained models, the standard Transformer shows the worst prediction-side coherence, while ARIMA performs best among statistical models. HierSTT achieves the best prediction-side HAgE overall, improving by approximately 13\% over ARIMA. On the right plot, which measures agreement between aggregated lower-level predictions and ground-truth higher-level values, HierSTT again outperforms all models, including Na\"ive and reconciliation baselines, reducing ground-truth HAgE by approximately 18\% relative to Seq2seq-LSTM, the second-best. This shows that its lower-level forecasts remain coherent and accurate when aggregated, unlike reconciliation baselines whose performance depends on base-forecast quality. HierSTT also reduces mean WAPE by over 30\% relative to the strongest competitors (N-BEATS, Na\"ive, and FA-Transformer), placing it in the lower-left corner of both plots.

A model can appear competitive in Fig.~\ref{fig:pred_accuracy} while performing poorly in Fig.~\ref{fig:acc_vs_hage}. TFT is a clear example: strong national-level accuracy but poor hospital-level performance leads to high mean WAPE and weak coherence. This highlights the importance of jointly evaluating accuracy and coherence across all levels.

HierSTT is the only model that consistently performs strongly on both dimensions, combining the best mean WAPE with learned hierarchical coherence. Error bars show that it is the most stable deep learning model across seeds.

\section{Conclusion}
\label{sec:conclusion}

In this work, we proposed HierSTT, a hierarchical Transformer-based framework for coherent multi-level forecasting of Emergency Department demand. HierSTT jointly models temporal dynamics, heterogeneous covariates, and cross-level dependencies through top-down conditioning, that allows higher-level forecasts to inform lower-level estimates, and a coherence-aware training objective. We additionally introduced a nationwide Portuguese ED dataset covering 81 hospitals across 5 regional health administrations.

Experiments show that HierSTT consistently outperforms statistical, non-hierarchical deep learning, and reconciliation baselines across all hierarchy levels, achieving over 30\% average WAPE improvement over the strongest competitor. Unlike post-hoc reconciliation methods or independently trained models, HierSTT jointly achieves high accuracy and near-coherent forecasts. Results also highlight the importance of evaluating hierarchical forecasting using both accuracy and coherence metrics such as HAgE.


\section{Acknowledgments}
This work was supported by Fundação para a Ciência e a Tecnologia (FCT)
through LARSyS funding
(DOIs:
\href{https://doi.org/10.54499/LA/P/0083/2020}
{\nolinkurl{10.54499/LA/P/0083/2020}},
\href{https://doi.org/10.54499/UIDP/50009/2020}
{\nolinkurl{10.54499/UIDP/50009/2020}}, and
\href{https://doi.org/10.54499/UIDB/50009/2020}
{\nolinkurl{10.54499/UIDB/50009/2020}}),
and PhD grant 2025.03757.BD
(DOI:
\href{https://doi.org/10.54499/2025.03757.BD}
{\nolinkurl{10.54499/2025.03757.BD}}).

%
%
%
\bibliographystyle{splncs04}
\bibliography{mybibfile}

@article{xu2021fm,
  title={FM-GRU: A time series prediction method for water quality based on seq2seq framework},
  author={Xu, Jianlong and Wang, Kun and Lin, Che and Xiao, Lianghong and Huang, Xingshan and Zhang, Yufeng},
  journal={Water},
  volume={13},
  number={8},
  pages={1031},
  year={2021},
  publisher={Multidisciplinary Digital Publishing Institute}
}

@article{masood2022multi,
  title={A multi-step time-series clustering-based Seq2Seq LSTM learning for a single household electricity load forecasting},
  author={Masood, Zaki and Gantassi, Rahma and Choi, Yonghoon},
  journal={Energies},
  volume={15},
  number={7},
  pages={2623},
  year={2022},
  publisher={MDPI}
}

@article{oreshkin2019n,
  title={N-BEATS: Neural basis expansion analysis for interpretable time series forecasting. arXiv 2019},
  author={Oreshkin, Boris N and Carpov, Dmitri and Chapados, Nicolas and Bengio, Yoshua},
  journal={arXiv preprint arXiv:1905.10437},
  year={2019}
}

@article{hyndman2002state,
  title={A state space framework for automatic forecasting using exponential smoothing methods},
  author={Hyndman, Rob J and Koehler, Anne B and Snyder, Ralph D and Grose, Simone},
  journal={International Journal of forecasting},
  volume={18},
  number={3},
  pages={439--454},
  year={2002},
  publisher={Elsevier}
}

@article{ji2024spatio,
  title={Spatio-temporal transformer network for weather forecasting},
  author={Ji, Junzhong and He, Jing and Lei, Minglong and Wang, Muhua and Tang, Wei},
  journal={IEEE Transactions on Big Data},
  volume={11},
  number={2},
  pages={372--387},
  year={2024},
  publisher={IEEE}
}

@article{ming2026transformers,
  title={Transformers Outperform Traditional Forecasting Models and Perform Comparably to Recurrent Neural Networks in the Prediction of Emergency Department Visits using Calendar and Meteorological Data},
  author={Ming, Chua and Leung, KH Benjamin and Shen, Yuzeng and Ho, Andrew FW},
  journal={Artificial Intelligence in Emergency Medicine},
  volume={1},
  pages={100006},
  year={2026},
  publisher={Elsevier}
}

@inproceedings{zhang2018base,
  title={Base on ETS model for Forcasting Emergency Department Visits},
  author={Zhang, Qingyu and Wang, Kang and Guo, Hainan and Yang, Shimiao and Li, Cui},
  booktitle={2018 IEEE 3rd Advanced Information Technology, Electronic and Automation Control Conference (IAEAC)},
  pages={2148--2151},
  year={2018},
  organization={IEEE}
}

@article{ming2025deep,
  title={Deep learning modelling to forecast emergency department visits using calendar, meteorological, internet search data and stock market price},
  author={Ming, Chua and Lee, Geraldine JW and Teo, Yao Neng and Teo, Yao Hao and Zhou, Xinyan and Ho, Elizabeth SY and Toh, Emma MS and Ong, Marcus Eng Hock and Tan, Benjamin YQ and Ho, Andrew FW},
  journal={Computer methods and programs in biomedicine},
  volume={267},
  pages={108808},
  year={2025},
  publisher={Elsevier}
}

@article{silva2023predicting,
  title={Predicting hospital emergency department visits accurately: A systematic review},
  author={Silva, Eduardo and Pereira, Margarida F and Vieira, Joana T and Ferreira-Coimbra, Jo{\~a}o and Henriques, Mariana and Rodrigues, Nuno F},
  journal={The International Journal of Health Planning and Management},
  volume={38},
  number={4},
  pages={904--917},
  year={2023},
  publisher={Wiley Online Library}
}

@article{turner2020effects,
  title={The effects of unexpected changes in demand on the performance of emergency departments},
  author={Turner, Alex J and Anselmi, Laura and Lau, Yiu-Shing and Sutton, Matt},
  journal={Health Economics},
  volume={29},
  number={12},
  pages={1744--1763},
  year={2020},
  publisher={Wiley Online Library}
}

@article{savioli2022emergency,
  title={Emergency department overcrowding: understanding the factors to find corresponding solutions},
  author={Savioli, Gabriele and Ceresa, Iride Francesca and Gri, Nicole and Bavestrello Piccini, Gaia and Longhitano, Yaroslava and Zanza, Christian and Piccioni, Andrea and Esposito, Ciro and Ricevuti, Giovanni and Bressan, Maria Antonietta},
  journal={Journal of personalized medicine},
  volume={12},
  number={2},
  pages={279},
  year={2022},
  publisher={MDPI}
}

@misc{AQI,
  title = {Technical Assistance Document for the Reporting of Daily Air Quality – the Air Quality Index (AQI)},
  author = {United States Environmental Protection Agency},
}

@article{vaswani2017attention,
  title={Attention is all you need},
  author={Vaswani, Ashish and Shazeer, Noam and Parmar, Niki and Uszkoreit, Jakob and Jones, Llion and Gomez, Aidan N and Kaiser, {\L}ukasz and Polosukhin, Illia},
  journal={Advances in neural information processing systems},
  volume={30},
  year={2017}
}

@article{alvarez2024evaluating,
  title={Evaluating the impact of exogenous variables for patients forecasting in an Emergency Department using Attention Neural Networks},
  author={{\'A}lvarez-Chaves, Hugo and Maseda-Zurdo, Iv{\'a}n and Mu{\~n}oz, Pablo and R-Moreno, Mar{\'\i}a D},
  journal={Expert Systems with Applications},
  volume={240},
  pages={122496},
  year={2024},
  publisher={Elsevier}
}

@article{kadri2020rnn,
  title={Rnn-based deep-learning approach to forecasting hospital system demands: application to an emergency department},
  author={Kadri, Farid and Abdennbi, Kahina},
  journal={International Journal of Data Science},
  volume={5},
  number={1},
  pages={1--25},
  year={2020},
  publisher={Inderscience Publishers (IEL)}
}

@Article{145815,
TITLE = {Comparison of Long Short-Term Memory and Convolutional Neural Network Models for Emergency Department Patients’ Arrival Daily Forecasting},
JOURNAL = {Journal of Archives in Military Medicine},
PUBLISHER = {},
VOLUME = {12},
NUMBER = {1},
ISSN = {},
AUTHOR = {Moosavi Kashani, S. and Zargar Balaye Jame, S. and Markazi-Moghaddam, N. and Omrani Nava, A.},
YEAR = {2024},
PAGES = {e140888},
}

@article{batal2001predicting,
  title={Predicting patient visits to an urgent care clinic using calendar variables},
  author={Batal, Holly and Tench, Jeff and McMillan, Sean and Adams, Jill and Mehler, Phillip S},
  journal={Academic Emergency Medicine},
  volume={8},
  number={1},
  pages={48--53},
  year={2001},
  publisher={Wiley Online Library}
}

@article{findley1998new,
  title={New capabilities and methods of the X-12-ARIMA seasonal-adjustment program},
  author={Findley, David F and Monsell, Brian C and Bell, William R and Otto, Mark C and Chen, Bor-Chung},
  journal={Journal of Business \& Economic Statistics},
  volume={16},
  number={2},
  pages={127--152},
  year={1998},
  publisher={Taylor \& Francis}
}

@book{box2015time,
  title={Time series analysis: forecasting and control},
  author={Box, George EP and Jenkins, Gwilym M and Reinsel, Gregory C and Ljung, Greta M},
  year={2015},
  publisher={John Wiley \& Sons}
}

@article{yule1971method,
  title={On a method of investigating periodicities in disturbed series with special reference to Wolfer’s sunspot numbers},
  author={Yule, George Udny},
  journal={Statistical Papers of George Udny Yule},
  pages={389--420},
  year={1971},
  publisher={Hafner Press New York}
}

@article{KAYACAN20101784,
title = {Grey system theory-based models in time series prediction},
journal = {Expert Systems with Applications},
volume = {37},
number = {2},
pages = {1784-1789},
year = {2010},
issn = {0957-4174},
author = {Erdal Kayacan and Baris Ulutas and Okyay Kaynak},
}

@article{https://doi.org/10.1002/sim.4780071007,
author = {Milner, P. C.},
title = {Forecasting the demand on accident and emergency departments in health districts in the trent region},
journal = {Statistics in Medicine},
volume = {7},
number = {10},
pages = {1061-1072},
year = {1988}
}

@article{10.4103/JETS.JETS_42_19,
author = {Moreno, Atilio and Muñoz, Oscar},
year = {2019},
month = {11},
pages = {268-273},
title = {Application of Queuing Theory to Optimize the Triage Process in a Tertiary Emergency Care (“ER”) Department},
volume = {12},
journal = {Journal of Emergencies Trauma and Shock},
}

@article{Boyle358,
	author = {Justin Boyle and Melanie Jessup and Julia Crilly and David Green and James Lind and Marianne Wallis and Peter Miller and Gerard Fitzgerald},
	title = {Predicting emergency department admissions},
	volume = {29},
	number = {5},
	pages = {358--365},
	year = {2012},
	publisher = {British Association for Accident and Emergency Medicine},
	journal = {Emergency Medicine Journal}
}

@article{10.1080/01605682.2022.2118629,
author = {Makridakis, Spyros and Spiliotis, Evangelos and Assimakopoulos, Vassilis and Semenoglou, Artemios-Anargyros and Mulder, Gary and Nikolopoulos, Konstantinos},
year = {2022},
month = {09},
pages = {1-20},
title = {Statistical, machine learning and deep learning forecasting methods: Comparisons and ways forward},
volume = {74},
journal = {Journal of the Operational Research Society},
}

@misc{olivares2023hierarchicalforecast,
      title={HierarchicalForecast: A Reference Framework for Hierarchical Forecasting in Python}, 
      author={Kin G. Olivares and Federico Garza and David Luo and Cristian Challú and Max Mergenthaler and Souhaib Ben Taieb and Shanika L. Wickramasuriya and Artur Dubrawski},
      year={2023},
      eprint={2207.03517},
      archivePrefix={arXiv},
      primaryClass={stat.ML}
}

@book{hyndman2018forecasting,
  title={Forecasting: principles and practice},
  author={Hyndman, Rob J and Athanasopoulos, George},
  year={2018},
  publisher={OTexts}
}

@article{Nasios_2022,
   title={Blending gradient boosted trees and neural networks for point and probabilistic forecasting of hierarchical time series},
   volume={38},
   ISSN={0169-2070},
   number={4},
   journal={International Journal of Forecasting},
   publisher={Elsevier BV},
   author={Nasios, Ioannis and Vogklis, Konstantinos},
   year={2022},
   month=oct, pages={1448–1459} 
   }

@inproceedings{10.1145/3616855.3635806,
author = {Wang, Shiyu},
title = {NeuralReconciler for Hierarchical Time Series Forecasting},
year = {2024},
isbn = {9798400703713},
publisher = {Association for Computing Machinery},
address = {New York, NY, USA},
booktitle = {Proceedings of the 17th ACM International Conference on Web Search and Data Mining},
pages = {731–739},
numpages = {9},
location = {, Merida, Mexico, },
series = {WSDM '24}
}

@inproceedings{caldas2022temporal,
  title="A Temporal Fusion Transformer for Long-term Explainable Prediction of Emergency Department Overcrowding",
  author="Caldas, Francisco M and Soares, Cláudia",
  booktitle="Joint European Conference on Machine Learning and Knowledge Discovery in Databases",
  pages={71--88},
  year={2022},
  organization={Springer}
}

@mastersthesis{pulkkinen2020forecasting,
  title={Forecasting emergency department arrivals with neural networks},
  author={Pulkkinen, Eetu},
  type={{B.S.} thesis},
  year={2020}
}

@article{mancuso2021machine,
  title={A machine learning approach for forecasting hierarchical time series},
  author={Mancuso, Paolo and Piccialli, Veronica and Sudoso, Antonio M},
  journal={Expert Systems with Applications},
  volume={182},
  pages={115102},
  year={2021},
  publisher={Elsevier}
}

\newpage 
\appendix
\section{Analysis of the Hierarchical Portuguese ED Dataset}

\label{sec:EDA}
\subsection{Variable Availability Across Hierarchical Levels}
\begin{table}[h!]
    \centering
    \caption{Available variables at each hierarchical level of the dataset of ED activity in Portugal.}
    \label{tab:dataset_table}
    \rowcolors{1}{}{lightgray}
    \setlength{\tabcolsep}{4pt}
    \begin{tabular}{l|c c c}
    \textbf{Variable Type} & \textbf{Hospital Level}  & \textbf{Regional Level} & \textbf{National Level} \\
    \specialrule{.1em}{.1em}{.1em}
        Hospital ID & \color{ForestGreen}\ding{51} & \color{BrickRed}\ding{55} & \color{BrickRed}\ding{55}\\
        RHA & \color{ForestGreen}\ding{51} & \color{ForestGreen}\ding{51} & \color{BrickRed}\ding{55}\\
        Locality & \color{ForestGreen}\ding{51} & \color{BrickRed}\ding{55} & \color{BrickRed}\ding{55}\\ 
        ED Visits (M1) & \color{ForestGreen}\ding{51} & \color{ForestGreen}\ding{51} & \color{ForestGreen}\ding{51}\\
        Triage (M2-M8) & \color{ForestGreen}\ding{51} &\color{ForestGreen}\ding{51}  & \color{ForestGreen}\ding{51}\\
        Waiting Time & \color{BrickRed}\ding{55} & \color{ForestGreen}\ding{51} & \color{ForestGreen}\ding{51}\\ 
        Calendar (weekday/month) & \color{ForestGreen}\ding{51} & \color{ForestGreen}\ding{51} & \color{ForestGreen}\ding{51}\\ 
        Holidays & \color{ForestGreen}\ding{51} & \color{ForestGreen}\ding{51}  & \color{ForestGreen}\ding{51} \\ 
        Air Quality (AQI) & \color{BrickRed}\ding{55} & \color{ForestGreen}\ding{51} &\color{BrickRed}\ding{55} \\ 
        Temperature & \color{BrickRed}\ding{55} & \color{ForestGreen}\ding{51} & \color{BrickRed}\ding{55}\\ 
        Mortality & \color{BrickRed}\ding{55} & \color{BrickRed}\ding{55} & \color{ForestGreen}\ding{51}\\ 
        Open ED Indicator & \color{ForestGreen}\ding{51} & \color{ForestGreen}\ding{51} & \color{ForestGreen}\ding{51}\\ 
        ED Type / Access / Age & \color{ForestGreen}\ding{51} & \color{BrickRed}\ding{55} & \color{BrickRed}\ding{55}\\ 
    \end{tabular}
\end{table}

The forecasting dataset integrates information from the hospital, regional, and national levels of the Portuguese healthcare system. Since variable availability differs across levels, the resulting feature space is heterogeneous across the hierarchy. Table~\ref{tab:dataset_table} summarizes the variables available at each level.

Hospital level variables includes local operational and demographic information, such as hospital identifier, Regional Health Administration (RHA), locality, emergency department (ED) visits, triage indicators, calendar variables, holidays, open/closed status, ED type, patient access pathways, and admitted age ranges. 

Regional level variables include aggregated ED demand and triage indicators, waiting times, calendar variables, holidays, open/closed status, and environmental variables, including air quality index (AQI) and temperature. Environmental variables are included only at this level due to the limited availability of reliable hospital-specific measurements.

National level variables include aggregated ED visits, triage indicators, waiting time, calendar variables, holidays, open/closed status, and mortality, providing a system-wide view of healthcare demand.

This level-dependent variable availability reflects real-world data collection constraints and motivates a hierarchical forecasting framework capable of combining heterogeneous level-specific information while producing coherent predictions across the hierarchy.

\subsection{Temporal Patterns in Hospital-Level ED Visits}

\begin{figure}[!h]
    \centering
    \begin{subfigure}{0.49\textwidth}
        \centering
        \includegraphics[width=\textwidth]{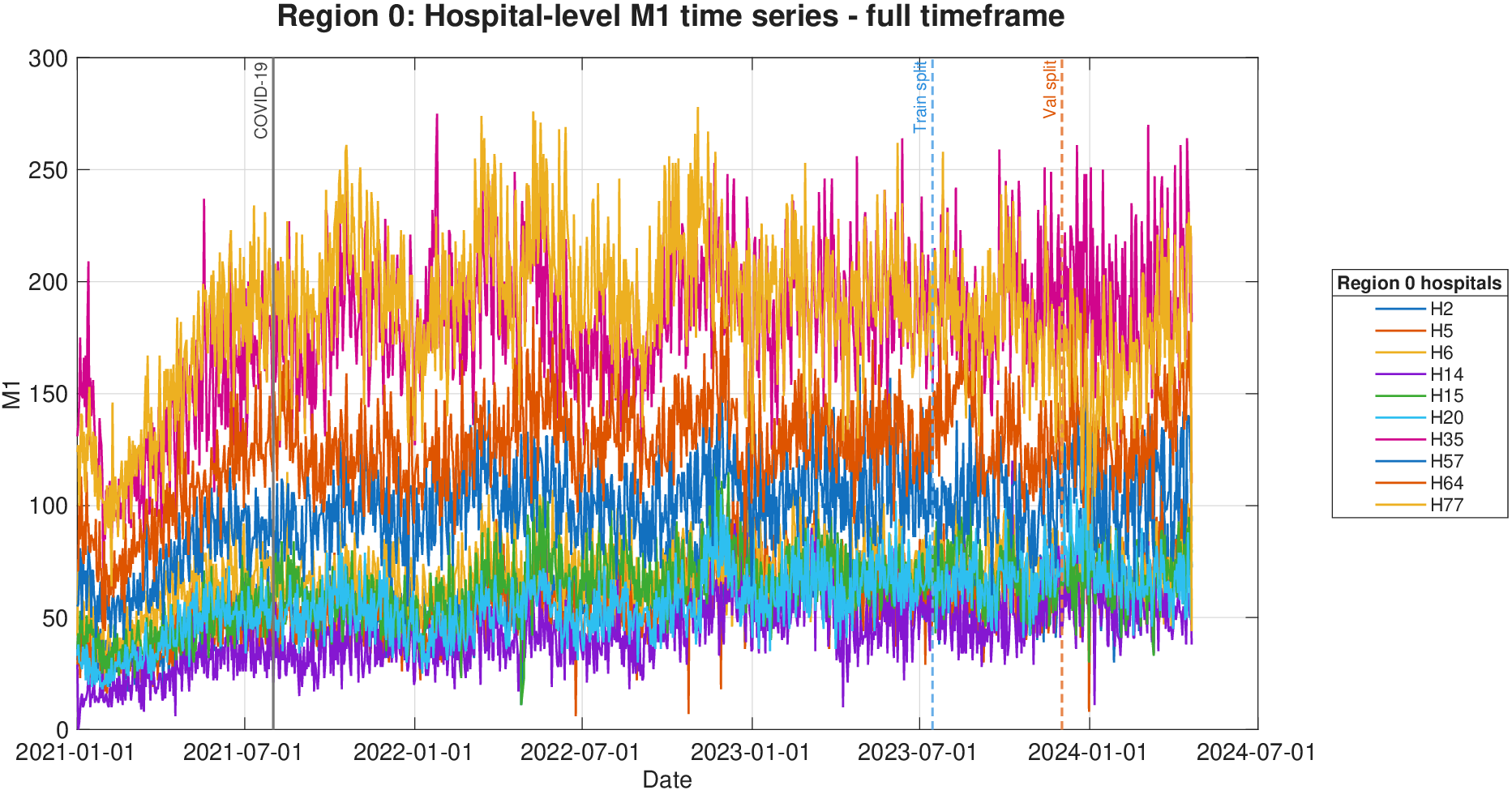}
        \caption{}
        \label{fig:dataset_reg_0}
    \end{subfigure}
    \hfill
    \begin{subfigure}{0.49\textwidth}
        \centering
        \includegraphics[width=\textwidth]{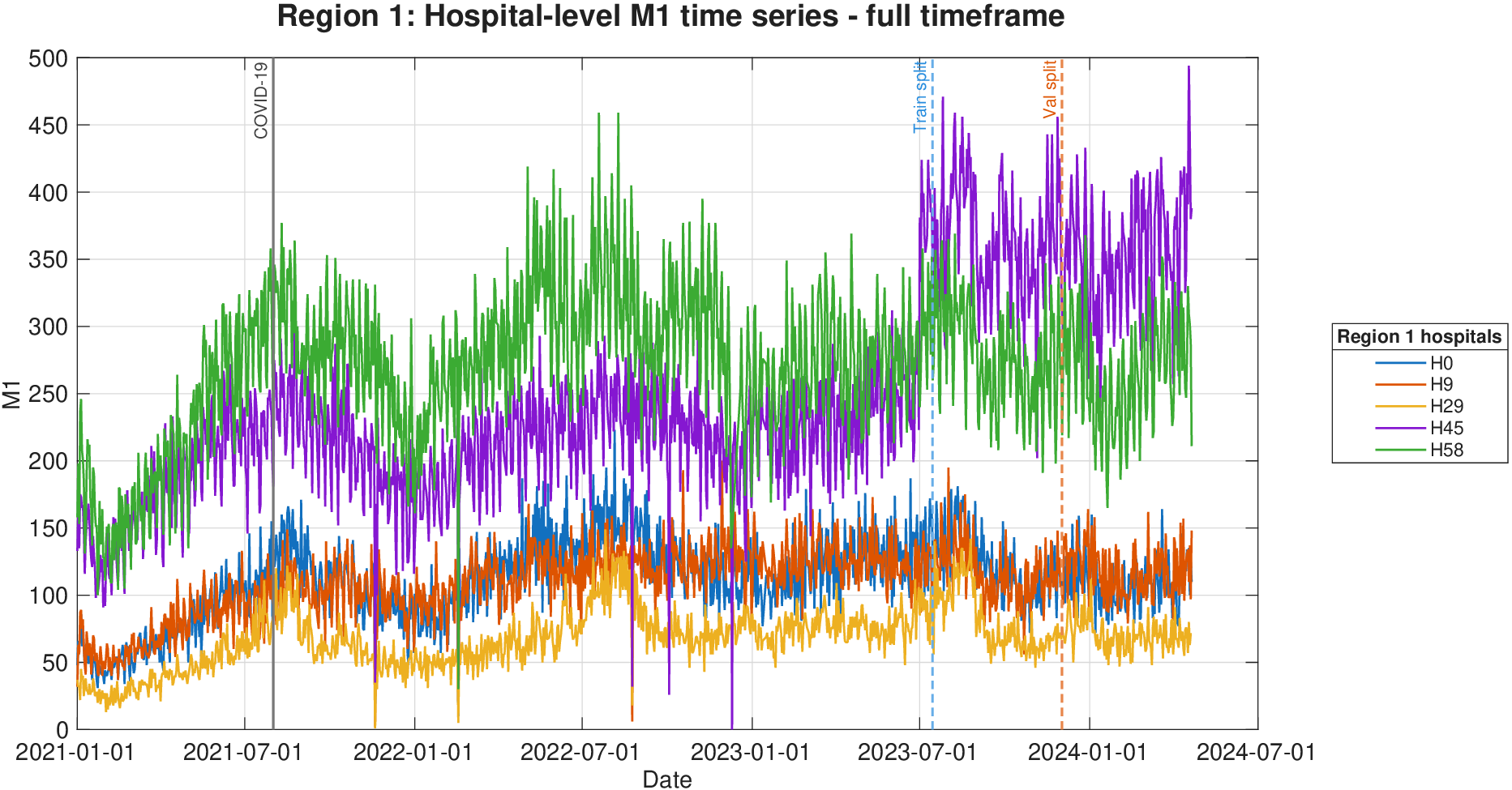}
        \caption{}
        \label{fig:dataset_reg_1}
    \end{subfigure}
    \hfill
    \begin{subfigure}{0.49\textwidth}
        \centering
        \includegraphics[width=\textwidth]{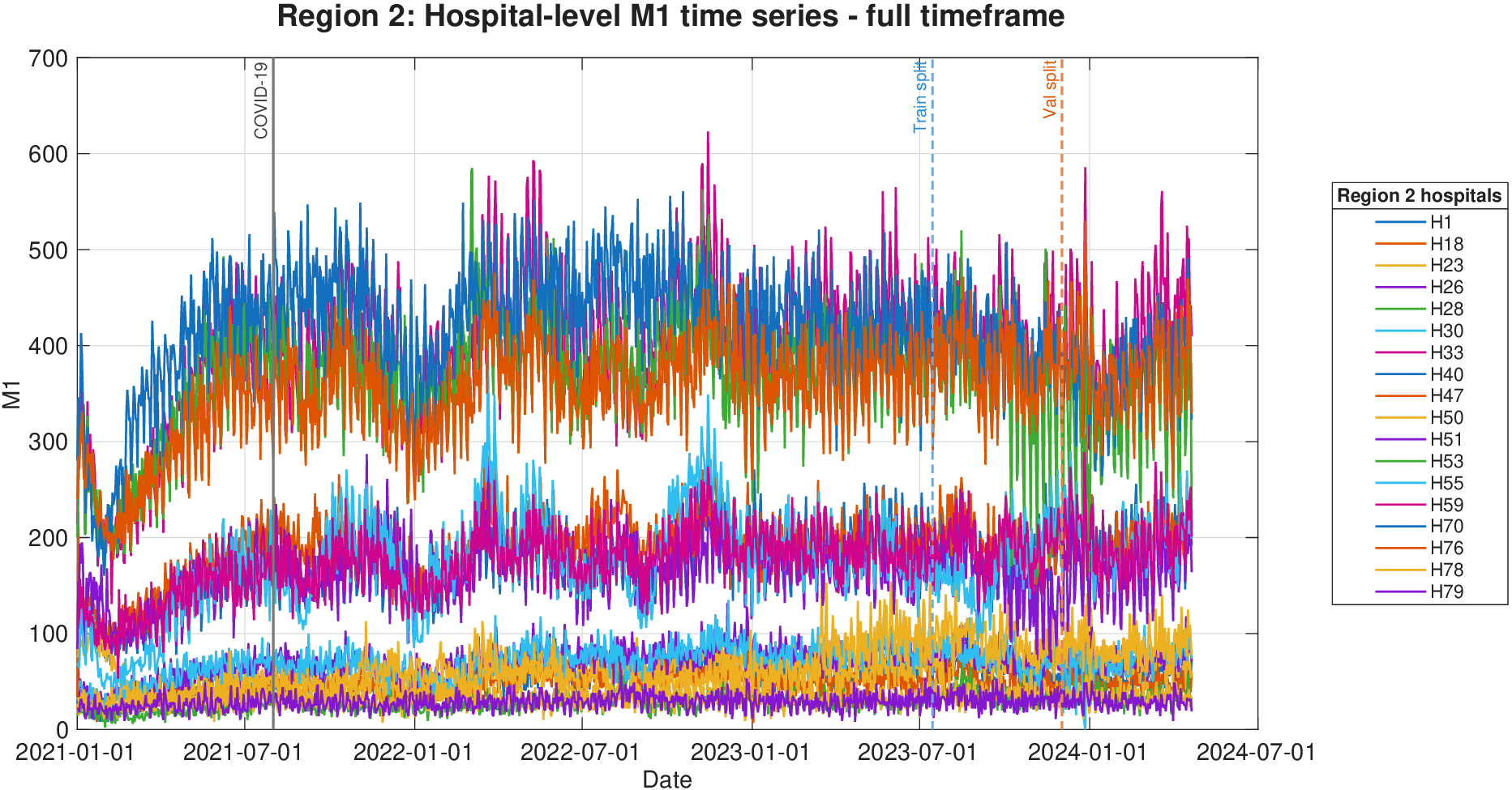}
        \caption{}
        \label{fig:dataset_reg_2}
    \end{subfigure}
    \hfill
    \begin{subfigure}{0.49\textwidth}
        \centering
        \includegraphics[width=\textwidth]{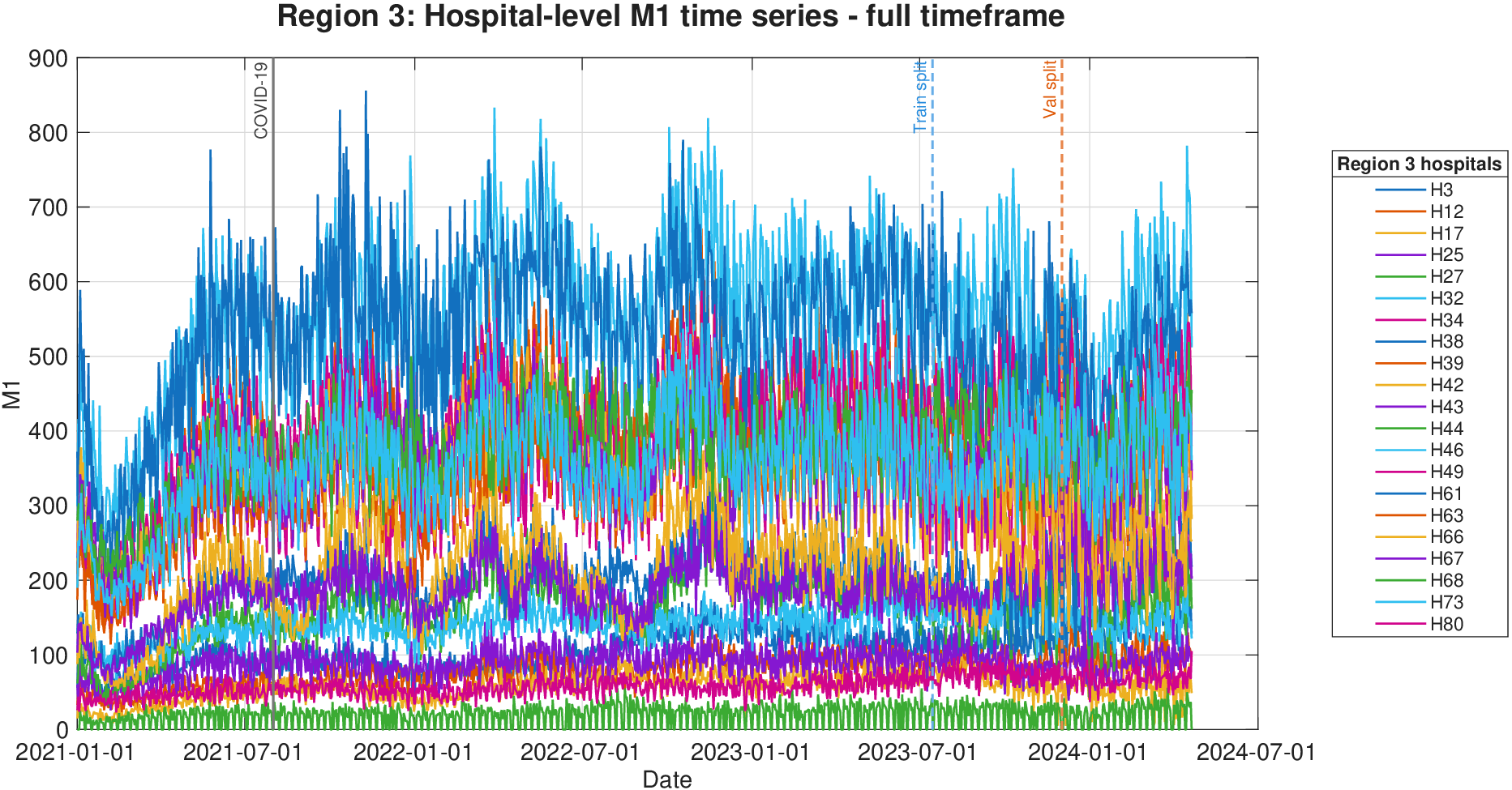}
        \caption{}
        \label{fig:dataset_reg_3}
    \end{subfigure}
    \hfill
    \begin{subfigure}{0.49\textwidth}
        \centering
        \includegraphics[width=\textwidth]{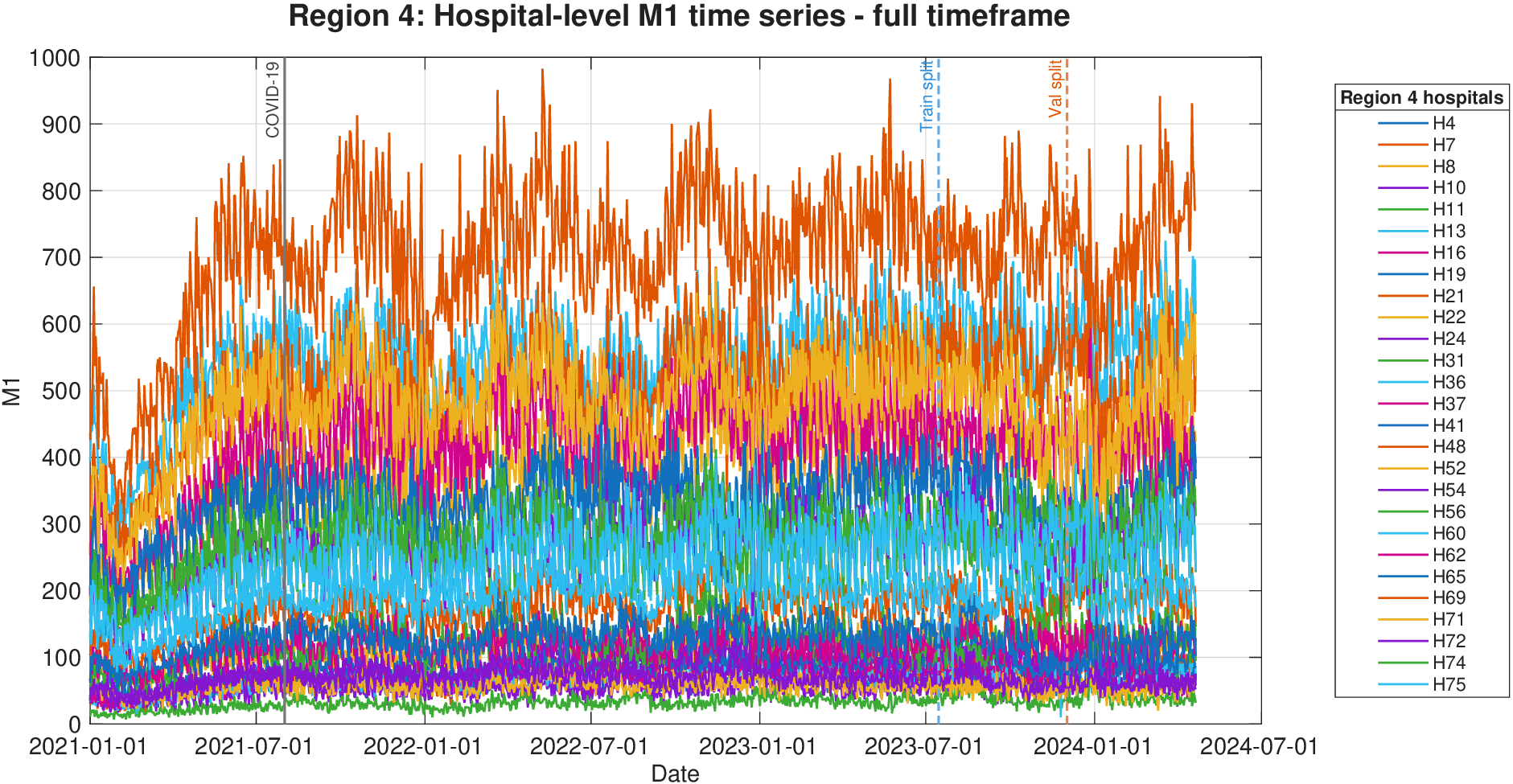}
        \caption{}
        \label{fig:dataset_reg_4}
    \end{subfigure}
        
    \caption{Daily hospital-level M1 time series within each region acrosss the full available time frame. Each line represents one hospital. The plots include the COVID-19 affected period and the subsequent modeling period.}
    \label{fig:dataset_M1s}
\end{figure}

To contextualize the forecasting task, Fig.~\ref{fig:dataset_M1s} shows the hospital-level M1 time series across the five regions over the full available time frame, including the COVID-19 period and the subsequent modeling period. The plots highlight the variability across hospitals, both in demand magnitude and temporal dynamics. This heterogeneity is particularly evident in Regions 2--4, where hospitals operate at different demand scales within the same regional unit.

The figures also show atypical demand patterns during the COVID-19 period, motivating the use of data from 1 August 2021 onward for forecasting experiments. The modeling start date and train/validation splits are indicated to illustrate the temporal evaluation protocol.

\begin{figure}[!h]
    \centering
    \begin{subfigure}{0.49\textwidth}
        \centering
        \includegraphics[width=\textwidth]{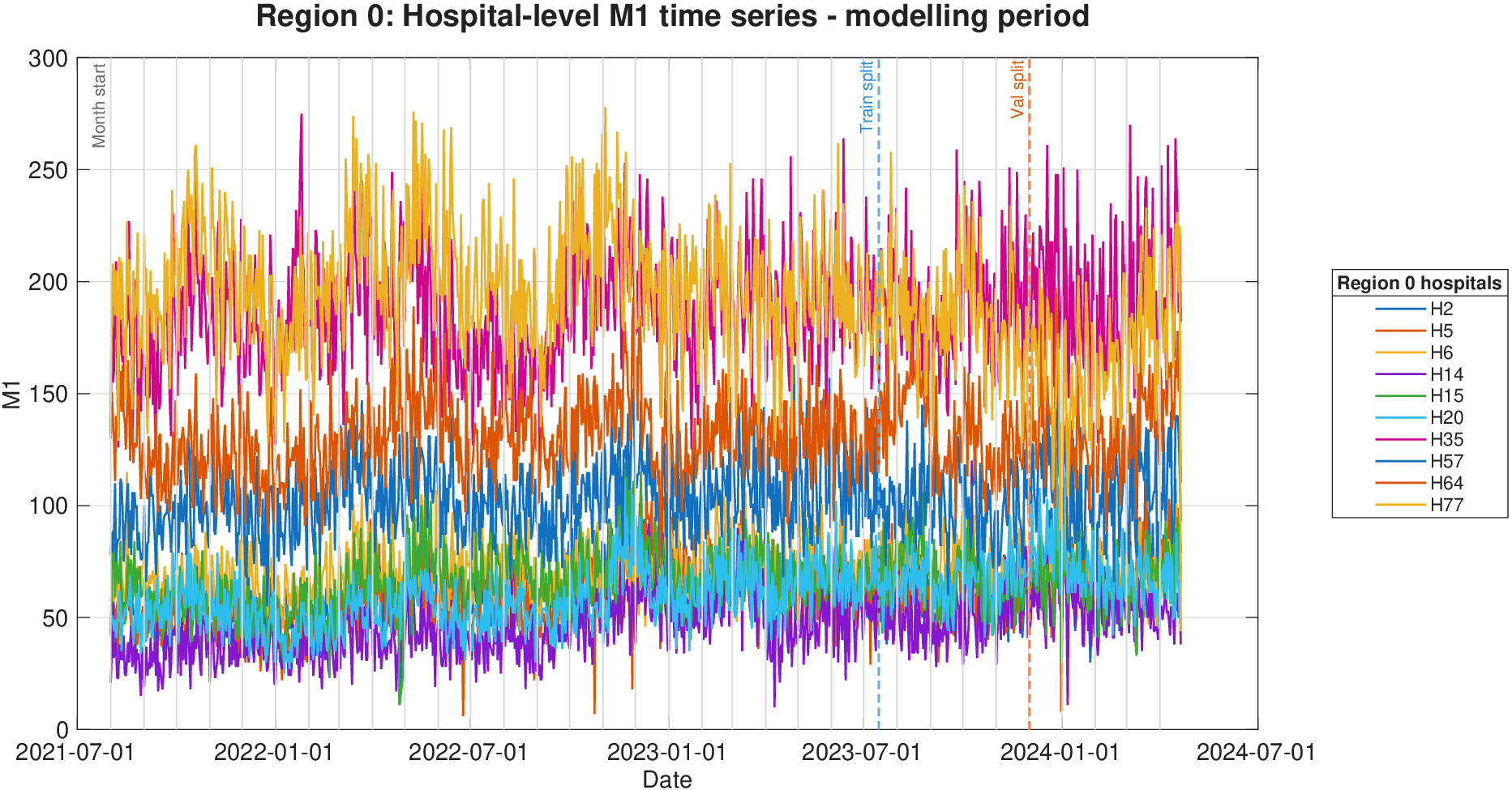}
        \caption{}
        \label{fig:reg0_model}
    \end{subfigure}
    \hfill
    \begin{subfigure}{0.49\textwidth}
        \centering
        \includegraphics[width=\textwidth]{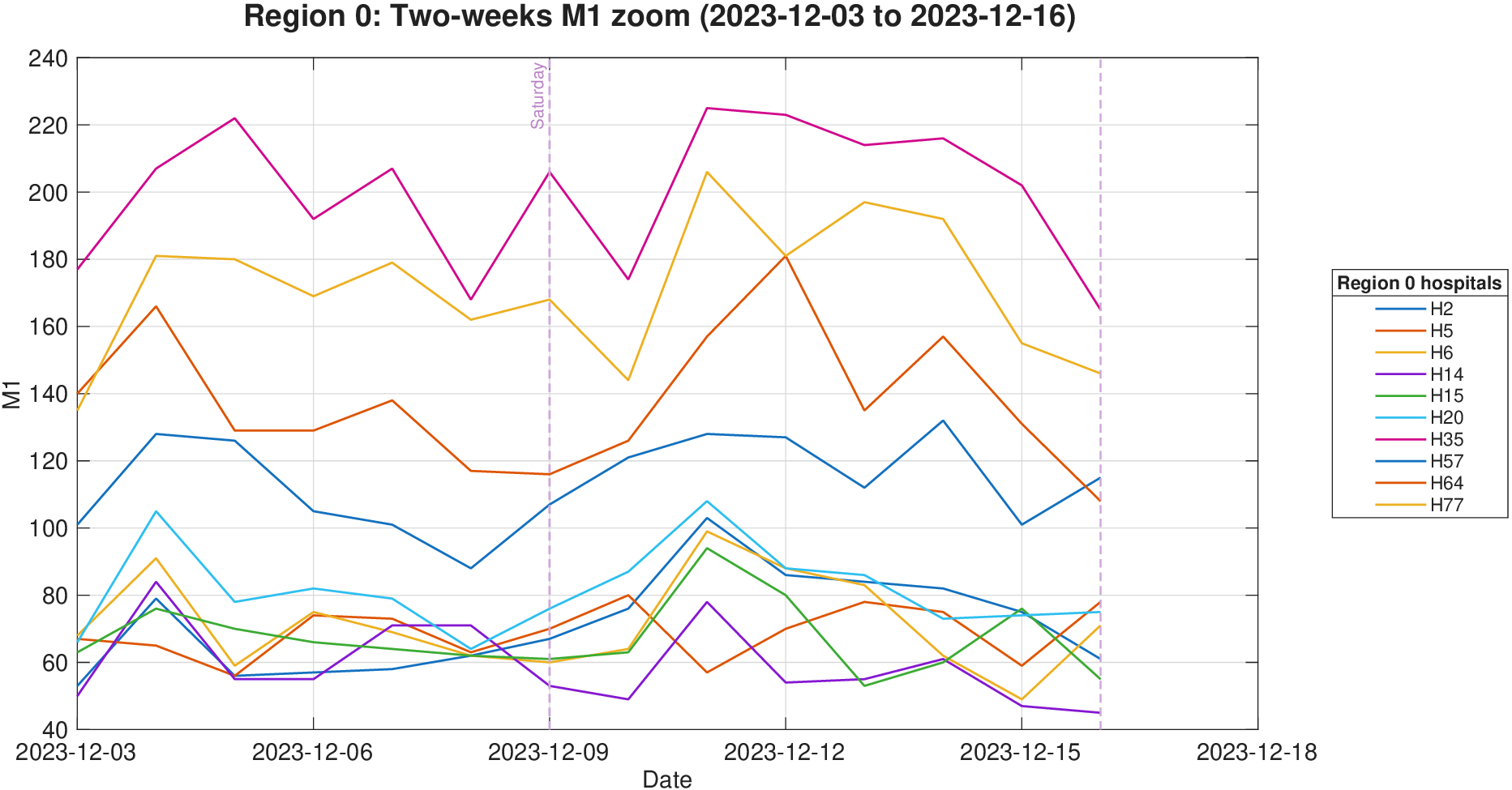}
        \caption{}
        \label{fig:reg0_zoom}
    \end{subfigure}
      \hfill
    \begin{subfigure}{0.49\textwidth}
        \centering
        \includegraphics[width=\textwidth]{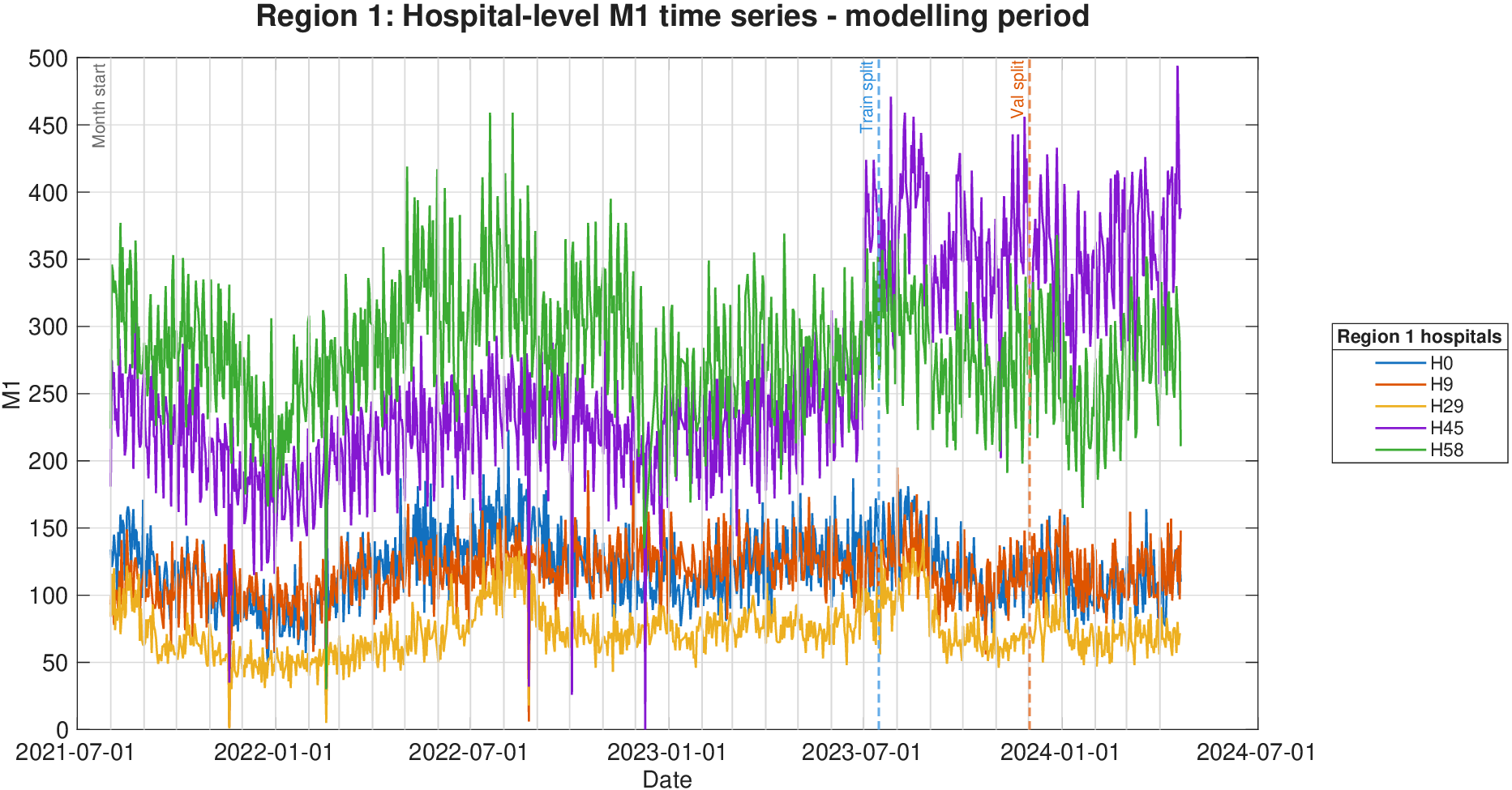}
        \caption{}
        \label{fig:reg1_model}
    \end{subfigure}
    \hfill
    \begin{subfigure}{0.49\textwidth}
        \centering
        \includegraphics[width=\textwidth]{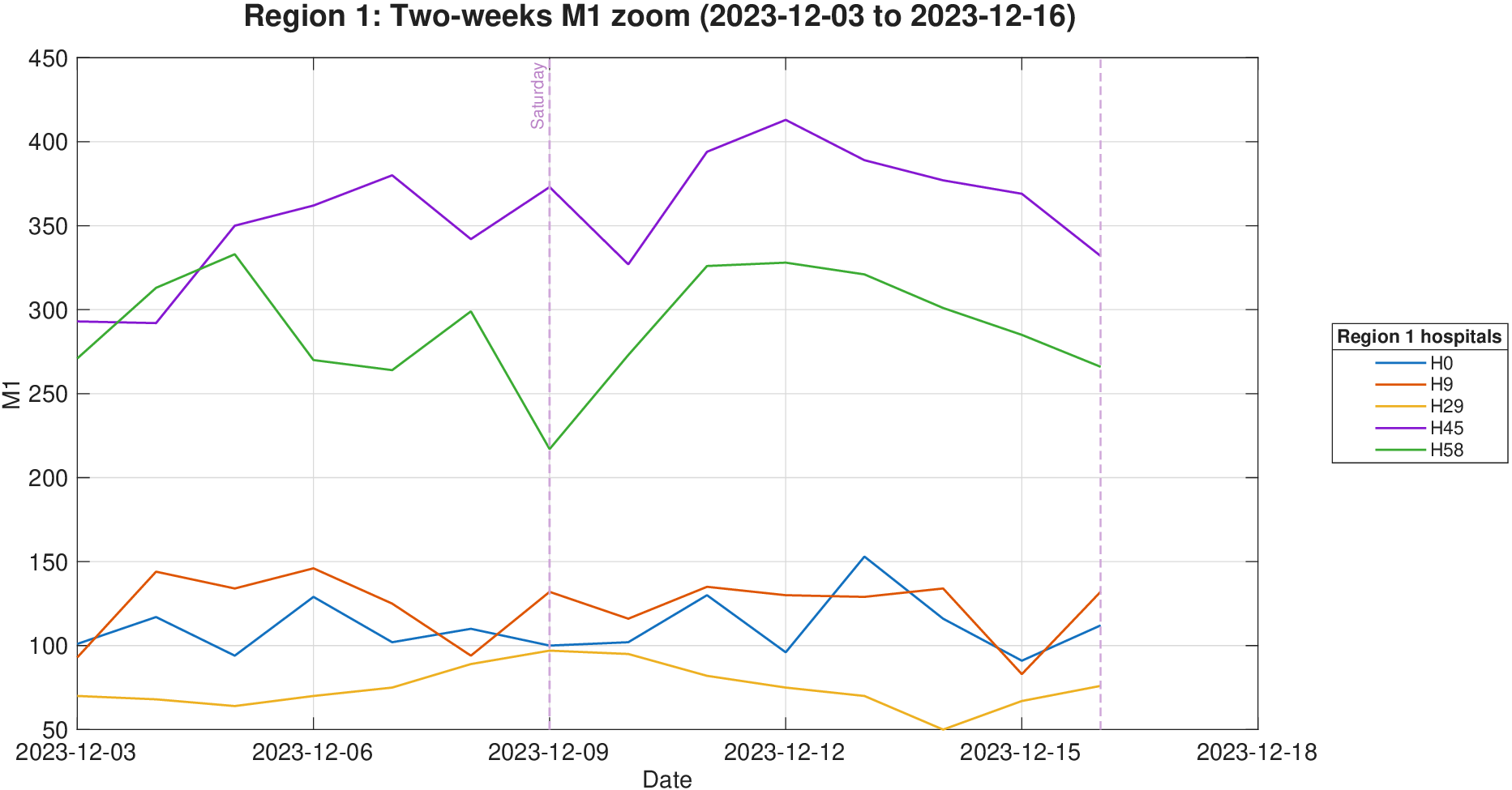}
        \caption{}
        \label{fig:reg1_zoom}
    \end{subfigure}
    
    \caption{Hospital-level M1 series over the modeling period for Region 0 (a) and Region 1 (c), selected as illustrative examples due to their smaller number of hospitals. Panels (b) and (d) show two-week zoomed examples of hospital-level trajectories for the same regions. Vertical dashed lines indicate Saturdays.}

    \label{fig:M1_zoom}
\end{figure}

Fig.~\ref{fig:M1_zoom} presents modeling-period and short-term examples for Regions 0 and 1, selected because their smaller number of hospitals facilitates visualization. The modeling-period plots reveal persistent differences between hospitals within the same region, while two-week zoom plots (Fig.~\ref{fig:reg0_zoom} and \ref{fig:reg1_zoom}) illustrate the short-term variability and recurring weekly patterns. The Saturday markers further highlight the importance of calendar structure, supporting the inclusion of future-known temporal covariates. 

\subsection{Sliding Window Sample Construction}

\begin{figure}[t]
    \centering
    \includegraphics[width=1\linewidth, trim={5mm 0 5mm 0},clip]{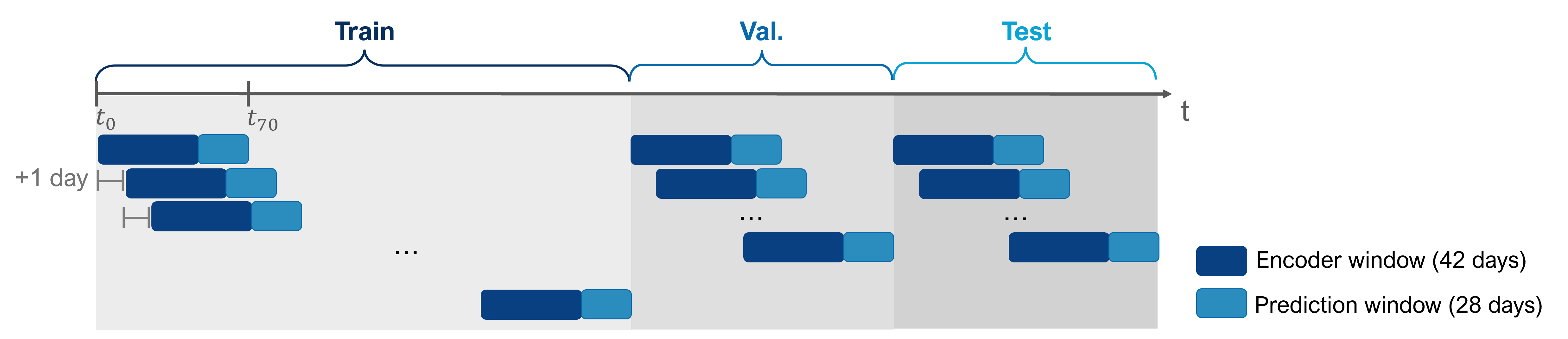}
    \caption{Sliding window strategy used to generate samples. Each sample consists of a 42-day encoder window (\(T_{in}\)) followed by a 28-day prediction window (\(T_{out}\)). Consecutive samples are shifted by one day.}
    \label{fig:data_window}
\end{figure}

Forecasting samples were generated using a sliding-window strategy with stride one. Each sample consists of a 42-day encoder window, \(T_{\text{in}}=42\), followed by a 28-day forecasting horizon, \(T_{\text{out}}=28\), as illustrated in Fig.~\ref{fig:data_window}. This input--output configuration was selected to capture short-term and weekly seasonal patterns while producing forecasts over a horizon relevant for operational planning in hospital emergency departments. 

The chronologically ordered time series were divided into training, validation, and test sets using cutoff dates of 15 July 2023 and 2 December 2023, resulting in 713 training days, 140 validation days, and 140 test days. No shuffling was applied, preserving the temporal order of the observations and preventing information leakage between splits.

Given a split with \(N\) consecutive days, the number of valid samples is computed as
\[
N - T_{\text{in}} - T_{\text{out}} + 1.
\]
Therefore, the 713 training days yield 644 training samples, while the 140-day validation and test periods each yield 71 samples.

\section{Ablation study: Effect of the Coherence Weight \(\alpha\)}
\label{sec:alpha_ablation}

\subsection{Loss Imbalance and the Role of \(\alpha\)}

The hierarchical loss defined in ~\eqref{eq:loss_general} supervises fundamentally different numbers of time series across its two terms. The accuracy term, \(\mathcal{L}^{(n)} + \mathcal{L}^{(r)} + \mathcal{L}^{(h)}\), operates over 87 individual series (81 hospital, 5 regional, and 1 national), whereas the coherence term \(\mathcal{L}_{coh}\) operates over 7 aggregation constrains: hospital-to-regional (5), hospital-to-national (1), and regional-to-national (1).

This asymmetry has a direct consequence on the effective influence of \(\alpha\). The per-constraint contribution of the coherence term is \(\alpha / 7\), while the per-series contribution of the accuracy term is \((1-\alpha) / 87\). The effective ratio between them is therefore:
\begin{equation}
    \frac{\alpha \cdot 87}{(1-\alpha) \cdot 7}\ ,
\end{equation}
which departs from the nominal ratio \(\alpha / (1-\alpha)\) implied by \(\alpha\) alone. At \(\alpha = 0.3\), the coherence term already exerts approximately five times more influence per constraint than the accuracy term does per series, meaning that coherence strongly shapes the learned representations even at low \(\alpha\) values. Conversely, even at \(\alpha = 0.999\), the effective ratio is approximately 12:1 rather than 999:1, confirming that the accuracy term is never truly suppressed and retains meaningful gradient signal throughout training. These considerations informed our choice of \(\alpha\) and highlight that its interpretation must account for the structural imbalance between the two loss components.

\subsection{Results Across \(\alpha\) values}

\begin{table*}[t]
\centering
\caption{Performance of HierSTT for different values of the coherence weight $\alpha$. HAgE denotes Hierarchical Aggregation Error computed using WAPE.}
\label{tab:alpha_ablation}
\resizebox{\textwidth}{!}{
\begin{tabular}{c c ccc c |c cc cc}
\toprule
  & & & & \multicolumn{2}{c|}{\textbf{WAPE [\%]}} & &\multicolumn{2}{c}{\textbf{Pred. HAgE [\%]}} & \multicolumn{2}{c}{\textbf{G.T. HAgE [\%]}} \\
  
\textbf{$\alpha$} & \textbf{Level} & \textbf{MAE} & \textbf{RMSE} & \textbf{Per Level} & \textbf{Avg.} & \textbf{Aggregation}&  \textbf{Per Agg.} & \textbf{ Avg.} &  \textbf{Per Agg.} & \textbf{ Avg.} \\

\hline
 \multicolumn{1}{>{\columncolor{lightgray}}l}{}& \cellcolor{lightgray}Hospital & \cellcolor{lightgray}\(22.16\pm0.42
\) & \cellcolor{lightgray}\(33.5 \pm 0.7
\) & \cellcolor{lightgray}\(10.87 \pm 0.20 \) & \multicolumn{1}{>{\columncolor{lightgray}}l|}{} & \cellcolor{lightgray}Hosp.$\to$Reg. & \cellcolor{lightgray}\(2.34 \pm 0.74\) &  \multicolumn{1}{>{\columncolor{lightgray}}l}{} & \cellcolor{lightgray}\(5.56 \pm 0.27\) & \multicolumn{1}{>{\columncolor{lightgray}}l}{} \\

  \multicolumn{1}{>{\columncolor{lightgray}}l}{} & \cellcolor{lightgray}Regional & \cellcolor{lightgray}\color{Green}\textbf{\(162.10 \pm 4.92\)} & \cellcolor{lightgray}\color{Green}\textbf{\(228.53 \pm 4.96
\)} & \cellcolor{lightgray}\color{Green}\textbf{\(4.90 \pm 0.15\)} & \multicolumn{1}{>{\columncolor{lightgray}}l|}{}  & \cellcolor{lightgray}Hosp.$\to$Nat. & \cellcolor{lightgray}\(3.28 \pm 0.65\) & \multicolumn{1}{>{\columncolor{lightgray}}l}{} & \cellcolor{lightgray}\(4.95 \pm 0.33\) & \multicolumn{1}{>{\columncolor{lightgray}}l}{} \\
  
 \multicolumn{1}{>{\columncolor{lightgray}}c}{\multirow{-3}{*}{\textbf{0}}} & \cellcolor{lightgray}National & \cellcolor{lightgray}\(929.89 \pm 133.19\) & \cellcolor{lightgray}\(1110.10 \pm 173.92\) & \cellcolor{lightgray}\(5.63 \pm 0.79\) & \multicolumn{1}{>{\columncolor{lightgray}}l|}{\multirow{-3}{*}{\underline{\(7.13 \pm 0.28\)}}}   & \cellcolor{lightgray}Reg.$\to$Nat. & \cellcolor{lightgray}\(3.49 \pm 0.86\) & \multicolumn{1}{>{\columncolor{lightgray}}l}{\multirow{-3}{*}{\(3.04 \pm 0.61 \)}}  & \cellcolor{lightgray}\(4.32 \pm 0.25\) & \multicolumn{1}{>{\columncolor{lightgray}}l}{\multirow{-3}{*}{\(4.94 \pm 0.25\)}}  \\
  
\hline
\multirow{3}{*}{\textbf{0.3}}
& Hospital & \color{Green}\textbf{\(22.03 \pm 0.18\)} & \color{Green}\textbf{\(33.22 \pm 0.37\)} & \color{Green}\textbf{\(10.81 \pm 0.09\)} & \multirow{3}{*}{\color{Green}\textbf{\(7.06 \pm 0.20\)}} & Hosp.$\to$Reg. & \(2.03 \pm 0.42\) & \multirow{3}{*}{\(2.83 \pm 0.53 \)} & \(5.56 \pm 0.16\) & \multirow{3}{*}{\underline{\(4.93 \pm 0.14\)}} \\
& Regional & \(162.77 \pm 4.55\) & \(229.70 \pm 5.99\) & \(4.92 \pm 0.14\) &  & Hosp.$\to$Nat. & \(3.11 \pm 0.65\) &  & \(4.93 \pm 0.19\) &  \\
& National & \color{Green}\textbf{\(897.78 \pm 114.69\)} & \color{Green}\textbf{\(1064.59 \pm 145.65\)} & \color{Green}\textbf{\(5.44 \pm 0.67\)} &  & Reg.$\to$Nat. &\(3.34 \pm 0.65\) &  & \color{Green}\textbf{\(4.30 \pm 0.17\)} &   \\

\hline
\multicolumn{1}{>{\columncolor{lightgray}}l}{} & \cellcolor{lightgray}Hospital & \cellcolor{lightgray}\(22.32 \pm 0.31\) & \cellcolor{lightgray}\(33.75 \pm 0.74\) & \cellcolor{lightgray}\(10.94 \pm 0.16\) & \multicolumn{1}{>{\columncolor{lightgray}}l|}{}  & \cellcolor{lightgray}Hosp.$\to$Reg. & \cellcolor{lightgray}\(2.10 \pm 0.62\) & \multicolumn{1}{>{\columncolor{lightgray}}l}{} & \cellcolor{lightgray}\color{Green}\textbf{\(5.45 \pm 0.25\)} & \multicolumn{1}{>{\columncolor{lightgray}}l}{} \\

\multicolumn{1}{>{\columncolor{lightgray}}l}{} & \cellcolor{lightgray}Regional & \cellcolor{lightgray}\(164.73 \pm 5.05
\) & \cellcolor{lightgray}\(230.73 \pm 4.98\) &  \cellcolor{lightgray}\(4.98 \pm 0.16\) & \multicolumn{1}{>{\columncolor{lightgray}}l|}{} & \cellcolor{lightgray}Hosp.$\to$Nat. & \cellcolor{lightgray}\(3.20 \pm 0.71\) &  \multicolumn{1}{>{\columncolor{lightgray}}l}{}& \cellcolor{lightgray}\color{Green}\textbf{\(4.84 \pm 0.33\)} & \multicolumn{1}{>{\columncolor{lightgray}}l}{} \\

\multicolumn{1}{>{\columncolor{lightgray}}c}{\multirow{-3}{*}{\textbf{0.5}}} & \cellcolor{lightgray}National & \cellcolor{lightgray}\(942.28 \pm 101.66\)& \cellcolor{lightgray}\(1126.85 \pm 117.78\) & \cellcolor{lightgray}\(5.72 \pm 0.61\) &  \multicolumn{1}{>{\columncolor{lightgray}}l|}{\multirow{-3}{*}{\(7.22 \pm 0.17 \)}}  & \cellcolor{lightgray}Reg.$\to$Nat. & \cellcolor{lightgray}\(3.49 \pm 0.48\) & \multicolumn{1}{>{\columncolor{lightgray}}l}{\multirow{-3}{*}{\(2.93 \pm 0.40 \)}} & \cellcolor{lightgray}\(4.43 \pm 0.21\) & \multicolumn{1}{>{\columncolor{lightgray}}l}{\multirow{-3}{*}{\color{Green}\textbf{\(4.90 \pm 0.23 \)}}} \\

\hline
\multirow{3}{*}{\textbf{0.7}}
& Hospital & \(22.18 \pm 0.26\) & \(33.42 \pm 0.49\) & \(10.88 \pm 0.12\) & \multirow{3}{*}{\(7.19 \pm 0.32 \)} & Hosp.$\to$Reg. & \(1.91 \pm 0.50 \) & \multirow{3}{*}{\(2.91 \pm 0.54\)} & \(5.53 \pm 0.23\) & \multirow{3}{*}{\(4.97 \pm 0.22 \)} \\
& Regional & \(164.46 \pm 6.70\) & \(231.54 \pm 7.60\) & \(4.97 \pm 0.21 \) &  & Hosp.$\to$Nat. & \(3.38 \pm0.87\) &  & \(4.96 \pm 0.28\) &  \\
& National & \(941.08 \pm 145.79\) & \(1131.79 \pm 206.36\) & \(5.71 \pm 0.90 \) &  & Reg.$\to$Nat. & \(3.43 \pm 0.83\) &  & \(4.42 \pm 0.27\) &  \\

\hline
\multicolumn{1}{>{\columncolor{lightgray}}l}{}
& \cellcolor{lightgray}Hospital & \cellcolor{lightgray}\(22.34 \pm 0.47\) & \cellcolor{lightgray}\(34.23 \pm 0.87\) & \cellcolor{lightgray}\(10.96 \pm 0.23\) & \multicolumn{1}{>{\columncolor{lightgray}}l|}{} & \cellcolor{lightgray}Hosp.$\to$Reg. & \cellcolor{lightgray}\(1.24 \pm 0.44 \) & \multicolumn{1}{>{\columncolor{lightgray}}l}{} & \cellcolor{lightgray}\(5.72 \pm 0.13\) & \multicolumn{1}{>{\columncolor{lightgray}}l}{}\\

\multicolumn{1}{>{\columncolor{lightgray}}l}{}& \cellcolor{lightgray}Regional & \cellcolor{lightgray}\(186.68 \pm 10.05\)& \cellcolor{lightgray}\(260.83 \pm 13.53\) & \cellcolor{lightgray}\(5.66 \pm 0.31 \) & \multicolumn{1}{>{\columncolor{lightgray}}l|}{} & \cellcolor{lightgray}Hosp.$\to$Nat. & \cellcolor{lightgray}\(1.79 \pm 0.80 \) & \multicolumn{1}{>{\columncolor{lightgray}}l}{} & \cellcolor{lightgray}\(5.00 \pm 0.18\) &  \multicolumn{1}{>{\columncolor{lightgray}}l}{}\\

\multicolumn{1}{>{\columncolor{lightgray}}c}{\multirow{-3}{*}{\textbf{0.99}}}& \cellcolor{lightgray}National & \cellcolor{lightgray}\(948.26 \pm 72.06\) & \cellcolor{lightgray}\(1134.55 \pm 89.38\) & \cellcolor{lightgray}\(5.76 \pm 0.43 \) & \multicolumn{1}{>{\columncolor{lightgray}}l|}{\multirow{-3}{*}{\(7.46 \pm 0.22 \)}} & \cellcolor{lightgray}Reg.$\to$Nat. & \cellcolor{lightgray}\(1.35 \pm 0.70 \) & \multicolumn{1}{>{\columncolor{lightgray}}l}{\multirow{-3}{*}{\underline{\(1.46 \pm 0.64\)}}}  & \cellcolor{lightgray}\(5.08 \pm 0.31\) &   \multicolumn{1}{>{\columncolor{lightgray}}l}{\multirow{-3}{*}{\(5.26 \pm 0.16\)}} \\

\hline
\multirow{3}{*}{\textbf{0.999}}
& Hospital & \(24.15 \pm 0.61\) & \(37.07 \pm 0.93\) & \(11.86 \pm 0.31\) & \multirow{3}{*}{\(7.74 \pm 0.23 \)} & Hosp.$\to$Reg. & \color{Green}\textbf{\(0.50 \pm 0.16 \)} & \multirow{3}{*}{\color{Green}\textbf{\(0.65 \pm 0.14\)}} & \(6.23 \pm 0.25\) & \multirow{3}{*}{\(5.66 \pm 0.20 \)} \\
& Regional & \(194.12 \pm 7.05\) & \(276.42 \pm 14.60\) & \(5.90 \pm 0.22 \) &  & Hosp.$\to$Nat. & \color{Green}\textbf{\(0.73 \pm 0.16\)} &  & \(5.41 \pm 0.20\) &  \\
& National & \(899.54 \pm 29.60\) & \(1076.37 \pm 47.72\) & \(5.47 \pm 0.19 \) &  & Reg.$\to$Nat. &\color{Green}\textbf{\(0.72 \pm 0.17\)} &  & \(5.33 \pm 0.16\) &  \\

\bottomrule
\end{tabular}
}
\end{table*}

Table~\ref{tab:alpha_ablation} reports the performance of HierSTT across different values of \(\alpha\), averaged over six fixed seeds (mean \(\pm\) standard deviation). Each configuration is evaluated on forecasting accuracy (MAE, RMSE, WAPE per level and average WAPE) and hierarchical coherence (prediction-side HAgE and ground-truth HAgE across the three aggregation transitions).

Several trends emerge. First, \(\alpha = 0.3\) achieves the lowest average WAPE (\(7.06 \pm 0.20\%\)) and the best hospital and national level MAE, RMSE, and WAPE, making it the best overall operating point. Second, prediction-side HAgE decreases as \(\alpha\) increases, from \(3.04\%\) at \(\alpha = 0\) to \(0.65\%\) at \(\alpha = 0.999\), confirming that the coherence objective progressively suppresses cross-level inconsistencies. Third, the accuracy-coherence trade-off remains mild up to \(\alpha = 0.5\) (average WAPE of \(7.22\%\)), but deteriorates for larger values: at \(\alpha = 0.99\), regional WAPE increases to \(5.66\%\) and at \(\alpha=0.999\), hospital WAPE rises to \(11.86\%\). This behavior is consistent with imbalance described in the previous subsection, where larger \(\alpha\) values increase the relative influence of the 7 aggregation constraints with respect to the 87 directly supervised time series.

Notably, even at \(\alpha = 0.999\), the model does not collapse to a degenerate solution. National-level WAPE remains at \(5.47\%\), only slightly above the best value of \(5.44\%\) obtained at \(\alpha = 0.3\). This behavior is also consistent with the loss-imbalance analysis above: although \(\alpha = 0.999\) appears to almost entirely suppress the accuracy term in nominal weight, the effective per-series contribution of the accuracy loss remains meaningful once the imbalance between the 87 supervised time series and the 7 aggregation constraints is taken into account. Consequently, the model continues to receive direct forecasting supervision, while the coherence term strongly reduces cross-level inconsistencies.

Based on these results, \(\alpha = 0.3\) is selected for all main experiments, as it provides the best trade-off between forecasting accuracy and hierarchical coherence.

\section{Extended Experimental Results}
\label{sec:results_detail}

\subsection{Overall Forecasting Performance}

\begin{table}[t]
\centering
\caption{Forecasting performance across all models and hierarchy levels. \textcolor{Green}{Green} = best, \underline{underline} = second best per column. FA-Transformer stands for Future-Aware Transformer, since the model uses future covariates in the decoder.}
\label{tab:main_results}
\resizebox{\textwidth}{!}{
\begin{tabular}{lll ccc ccc ccc}
\toprule
& & &\multicolumn{3}{c}{\textbf{Hospital}} & \multicolumn{3}{c}{\textbf{Regional}} & \multicolumn{3}{c}{\textbf{National}} \\
\cmidrule(lr){4-6} \cmidrule(lr){7-9} \cmidrule(lr){10-12}
 & \multicolumn{2}{c}{\textbf{Model}} & MAE & RMSE & WAPE [\%]  & MAE & RMSE & WAPE [\%] & MAE & RMSE & WAPE [\%] \\
 
\midrule

\multicolumn{1}{>{\columncolor{white}}l}{} & \cellcolor{lightgray}Na\"ive & \cellcolor{lightgray}       & \cellcolor{lightgray}25.5 & \cellcolor{lightgray}31.6 & \cellcolor{lightgray}15.4 & \cellcolor{lightgray}\underline{261.6}  & \cellcolor{lightgray}\underline{313.0} & \cellcolor{lightgray}\underline{8.1} & \cellcolor{lightgray}1251.1 & \cellcolor{lightgray}1461.7 & \cellcolor{lightgray}7.6 \\

 & ARIMA & & 28.1 & 33.8 & 15.6 & 345.3  & 411.0 & 10.4 & 1721.7 & 2034.5 & 10.5 \\

\multicolumn{1}{>{\columncolor{white}}l}{} & \cellcolor{lightgray}ETS & \cellcolor{lightgray}  & \cellcolor{lightgray}29.9 & \cellcolor{lightgray}35.9 & \cellcolor{lightgray}16.5 & \cellcolor{lightgray}362.0  & \cellcolor{lightgray}431.1 & \cellcolor{lightgray}11.2  & \cellcolor{lightgray}1879.4 & \cellcolor{lightgray}2202.7 & \cellcolor{lightgray}11.5 \\
 
\arrayrulecolor{gray}\cline{2-12}
& LSTM  &  & \(26.4\pm2.7\) & \(32.7 \pm 3.0\) & \(14.7 \pm 1.0\) & \(359.9 \pm 117.0\)  & \(423.2 \pm 127.9\) & \(12.4 \pm 3.1\) & \(	1360.4 \pm 188.5\) & \(1708.2 \pm 215.0\) & \(8.2 \pm 1.1\)\\

\multicolumn{1}{>{\columncolor{white}}l}{} & \multicolumn{2}{l}{\cellcolor{lightgray}Seq2seq-LSTM}  & \cellcolor{lightgray}\(24.5 \pm 1.1\) & \cellcolor{lightgray}\underline{\(	30.7 \pm 1.2\)} & \cellcolor{lightgray}\(14.1 \pm 0.7\) & \cellcolor{lightgray}\(299.2 \pm 132.9\)  & \cellcolor{lightgray}\(357.9 \pm 152.6\) & \cellcolor{lightgray}\(13.4 \pm 8.4\) & \cellcolor{lightgray}\(1054.2\pm 235.2\) & \cellcolor{lightgray}\(1312.5 \pm 281.3\) & \cellcolor{lightgray}\(6.4 \pm 1.4\)  \\
 
 & \multicolumn{2}{l}{Seq2seq-GRU}    & \(25.1 \pm 2.8\) & \(31.2\pm3.4\) & \(14.4 \pm 1.0\) & \(345.9 \pm 40.4\)  & \(422.8 \pm 50.3\) & \(15.0 \pm 6.6\) & \(948.2 \pm 102.9\)  & \(1187.7 \pm 128.2\) & \underline{\(5.7\pm0.6\)} \\

\multicolumn{1}{>{\columncolor{white}}l}{} & \multicolumn{2}{l}{\cellcolor{lightgray}Transformer}  & \cellcolor{lightgray}\(28.1 \pm 1.4\) & \cellcolor{lightgray}\(34.8 \pm 1.5\) & \cellcolor{lightgray}\(15.8 \pm 0.4\)  & \cellcolor{lightgray}\(	498.6 \pm 74.0\) & \cellcolor{lightgray}\(605.2 \pm 82.4\) & \cellcolor{lightgray}\(15.7 \pm 3.0\) & \cellcolor{lightgray}\(	3301.9 \pm 614.3\) & \cellcolor{lightgray}\(4358.4 \pm 738.4\) & \cellcolor{lightgray}\(20.0 \pm 3.7\) \\
 
 & \multicolumn{2}{l}{FA-Transformer} & \(26.9 \pm 0.9\) & \(33.3 \pm 1.0\) & \(14.7 \pm 0.4\)  & \(307.0 \pm 80.8\)  & \(354.4 \pm 86.7\) & \(10.3 \pm 2.7\) & \(1020.4 \pm 435.1\)  & \(1227.6 \pm 467.4\) & \(6.2 \pm 2.7\) \\

 \multicolumn{1}{>{\columncolor{white}}l}{}&  \cellcolor{lightgray}TFT & \cellcolor{lightgray} & \cellcolor{lightgray}\(82.0 \pm 10.8\) & \cellcolor{lightgray}\(89.8 \pm 10.5\) & \cellcolor{lightgray}\(46.2 \pm 10.0\) & \cellcolor{lightgray}\(381.3 \pm 28.7\) & \cellcolor{lightgray}\(455.0\pm29.6\) & \cellcolor{lightgray}\(12.1 \pm 1.7\)  & \cellcolor{lightgray}\color{Green}\textbf{\(896.3\pm119.3\)}  & \cellcolor{lightgray}\underline{\(1072.0 \pm 134.0\)} & \cellcolor{lightgray}\color{Green}\textbf{\(5.4 \pm 0.7\)} \\
 
\multicolumn{1}{>{\columncolor{white}}l}{\multirow{-11}{*}{\rotatebox[origin=c]{90}{\textbf{Non-Hierarchical}}}} & N-BEATS    &               & \underline{\(23.8 \pm 0.9\)} & \color{Green}\textbf{\(29.7 \pm 1.0\)} & \underline{\(14.1 \pm 0.6\)} & \(330.9 \pm 31.3\)  & \(408.3 \pm 35.4\) & \(10.6 \pm 1.3\)  & \(1149.8 \pm 166.5\) & \(1402.1 \pm 200.8\) & \(7.0 \pm 1.0\) \\
 \arrayrulecolor{black}
\midrule
 \multirow{10}{*}{\rotatebox[origin=c]{90}{\textbf{Hierarchical}}}
 &\multicolumn{1}{>{\columncolor{lightgray}}l}{}  & \cellcolor{lightgray}Na\"ive        & \cellcolor{lightgray}25.5 & \cellcolor{lightgray}31.6 & \cellcolor{lightgray}15.4 & \cellcolor{lightgray}\underline{261.6}  & \cellcolor{lightgray}\underline{313.0} & \cellcolor{lightgray}\underline{8.1} & \cellcolor{lightgray}1251.1 & \cellcolor{lightgray}1461.7 & \cellcolor{lightgray}7.6\\
  
 &\multicolumn{1}{>{\columncolor{lightgray}}l}{}  & ARIMA            & 30.0 & 35.9 & 16.4 & 363.1  & 430.2 & 10.9 &  1721.7 & 2034.5 & 10.5  \\
 
 &\multicolumn{1}{>{\columncolor{lightgray}}l}{\multirow{-3}{*}{Top-Down}}  & \cellcolor{lightgray}ETS              & \cellcolor{lightgray}32.7 & \cellcolor{lightgray}39.1 & \cellcolor{lightgray}17.7 & \cellcolor{lightgray}393.7  & \cellcolor{lightgray}466.4 & \cellcolor{lightgray}12.1 & \cellcolor{lightgray}1879.4 & \cellcolor{lightgray}2202.7 & \cellcolor{lightgray}11.5  \\

&\multirow{3}{*}{Middle-Out} & Na\"ive      & 25.5 & 31.6 & 15.4 & \underline{261.6}  & \underline{313.0} & \underline{8.1} & 1251.1 & 1461.7 & 7.6\\
  
 &  & \cellcolor{lightgray}ARIMA          & \cellcolor{lightgray}29.2 & \cellcolor{lightgray}35.0 & \cellcolor{lightgray}16.1 & \cellcolor{lightgray}345.3  & \cellcolor{lightgray}411.0 & \cellcolor{lightgray}10.4 & \cellcolor{lightgray}1612.6 & \cellcolor{lightgray}1910.2 & \cellcolor{lightgray}9.8  \\
 
 &  & ETS            & 31.3 & 37.6 & 17.1 & 362.0  & 431.1 & 11.2 & 1706.1 & 2013.2 & 10.4 \\
 
 & \multicolumn{1}{>{\columncolor{lightgray}}l}{} & \cellcolor{lightgray}Na\"ive       &  \cellcolor{lightgray}25.5 & \cellcolor{lightgray}31.6 & \cellcolor{lightgray}15.4 & \cellcolor{lightgray}\underline{261.6}  & \cellcolor{lightgray}\underline{313.0} & \cellcolor{lightgray}\underline{8.1} & \cellcolor{lightgray}1251.1 & \cellcolor{lightgray}1461.7 & \cellcolor{lightgray}7.6 \\
 
  & \multicolumn{1}{>{\columncolor{lightgray}}l}{}&ARIMA           & 28.1 & 33.8 & 15.6 &  321.0  & 381.7 & 9.8  &  1549.9 & 1830.1 & 9.4   \\
 
 & \multicolumn{1}{>{\columncolor{lightgray}}l}{\multirow{-3}{*}{Bottom-Up}} &\cellcolor{lightgray}ETS             & \cellcolor{lightgray}29.9 & \cellcolor{lightgray}35.9 & \cellcolor{lightgray}16.5 &  \cellcolor{lightgray}335.9  & \cellcolor{lightgray}398.3 & \cellcolor{lightgray}10.2 & \cellcolor{lightgray}1629.7 & \cellcolor{lightgray}1917.2 & \cellcolor{lightgray}9.9  \\
 
\cline{2-12}
& \multicolumn{2}{l}{\textbf{HierSTT (ours)}}& \color{Green}\textbf{\(22.0 \pm 0.2
\)} & \(33.2 \pm 0.4\) & \color{Green}\textbf{\(10.8 \pm 0.1\)}  & \color{Green}\textbf{\(162.8 \pm 4.6
\)} & \color{Green}\textbf{\(229.7 \pm 6.0
\)} & \color{Green}\textbf{\(4.9 \pm 0.1\)}  & \underline{\(897.8 \pm 114.7
\)} & \color{Green}\textbf{\(1064.6 \pm 145.7
\)} & \color{Green}\textbf{\(5.4 \pm 0.7\)}  \\
\bottomrule
\end{tabular}
}
\end{table}

Table~\ref{tab:main_results} reports MAE, RMSE and WAPE across all hierarchy levels. Classical statistical baselines remain competitive, particularly Na\"ive forecasting, but are generally outperformed by deep learning models at the hospital level, where ED demand exhibits greater variability and complexity, and at national level. Among the hierarchical reconciliation approaches (top-down, middle-out, bottom-up), reconciliation rarely improves over the corresponding non-hierarchical baselines. The N\"ive model yields identical results across all strategies because repeating the last observed 28-day preserves scale consistency by construction. The only exception is bottom-up reconciliation applied to ARIMA and ETS, which improves regional and national performance, suggesting that aggregating sufficiently structured lower-level forecasts can propagate useful signal upward. Overall, these results indicate that hierarchy alone is insufficient and effective cross-level integration during training is essential. 

Non-hierarchical deep learning models generally improve forecasting accuracy, although performance varies substantially across hierarchy levels. TFT achieves strong regional and national WAPE (12.1\% and 5.4\%, respectively) but performs poorly at the hospital level (WAPE 46.2\%), highlighting the limitations of single-level optimization. N-BEATS achieves the best hospital-level RMSE (29.7) and TFT attains the best national level MAE (896.3). Seq2seq architectures perform consistently across levels, though without leading at any.

HierSTT outperforms all baselines at the hospital and regional levels, reducing hospital WAPE by 23\% relative to N-BEATS (10.8\% vs.\ 14.1\%) and regional MAE by 38\% relative to the best non-hierarchical competitor (162.8 vs.\ 261.6). At the national level, it remains competitive with TFT. Notably, the TFT component of HierSTT preserve its standalone national level performance when trained jointly across all levels (MAE 897.8 vs.\ 896.3), showing that end-to-end hierarchical training does not compromise higher-level forecasting. Overall, these results suggest that jointly modeling spatial dependencies and hierarchical context improves performance across all aggregation levels without sacrificing accuracy at any single level.


\subsection{Hierarchical Coherence Analysis}

\begin{table*}[t]
\centering
\caption{HAgE across aggregation transitions. \textit{Pred.} measures inconsistency between predictions at different levels, while \textit{G.T.} measures the discrepancy between aggregated lower-level predictions and observed higher-level values. `\textcolor{blue}{\textbf{---}}' indicate reconciliation baselines where prediction-side HAgE is zero by construction. \textcolor{Green}{Green} = best, \underline{underline} = second best.}
\label{tab:hage_results}
\resizebox{0.7\textwidth}{!}{
\begin{tabular}{lll cc cc cc}
\toprule
& & & \multicolumn{2}{c}{\textbf{Hosp.\ $\to$ Reg.}} & \multicolumn{2}{c}{\textbf{Hosp.\ $\to$ Nat.}} & \multicolumn{2}{c}{\textbf{Reg.\ $\to$ Nat.}} \\
\cmidrule(lr){4-5} \cmidrule(lr){6-7} \cmidrule(lr){8-9}

 & \multicolumn{2}{c}{\textbf{Model}} & Pred.\ [\%] & G.T.\ [\%] & Pred.\ [\%] & G.T.\ [\%] & Pred.\ [\%] & G.T.\ [\%]  \\
 
\midrule

\multicolumn{1}{>{\columncolor{white}}l}{} & \cellcolor{lightgray}Na\"ive & \cellcolor{lightgray}       & \cellcolor{lightgray}\color{blue}\textbf{---}  & \cellcolor{lightgray}8.0 & \cellcolor{lightgray}\color{blue}\textbf{---} & \cellcolor{lightgray}7.6 & \cellcolor{lightgray}\color{blue}\textbf{---}    & \cellcolor{lightgray}\underline{7.6} \\

 & ARIMA & & \underline{3.7} & 9.7 & \underline{3.9}  & 9.4 & \color{Green}\textbf{2.1} & 9.8 \\

\multicolumn{1}{>{\columncolor{white}}l}{} & \cellcolor{lightgray}ETS & \cellcolor{lightgray}  & \cellcolor{lightgray}4.1 & \cellcolor{lightgray}10.2 & \cellcolor{lightgray}4.8 & \cellcolor{lightgray}9.9 & \cellcolor{lightgray}3.0  & \cellcolor{lightgray}10.4 \\
 
\arrayrulecolor{gray}\cline{2-9}
& LSTM  &  & \(	8.7 \pm 4.6\)  & \(8.2 \pm 1.6\)  & \(10.1 \pm 2.2\)  & \(6.7 \pm 1.2\)  & \(11.6 \pm 4.8\) & \(8.6 \pm 3.1\)\\

\multicolumn{1}{>{\columncolor{white}}l}{} & \multicolumn{2}{l}{\cellcolor{lightgray}Seq2seq-LSTM} & \cellcolor{lightgray}\(7.0 \pm 3.5\)& \cellcolor{lightgray}\(6.5 \pm 0.7\) & \cellcolor{lightgray}\(6.0 \pm 0.7\) & \cellcolor{lightgray}\underline{\(4.9 \pm 0.3\)}  & \cellcolor{lightgray}\(6.1 \pm 1.6\) & \cellcolor{lightgray}\(6.5 \pm 2.3\)  \\
 
 & \multicolumn{2}{l}{Seq2seq-GRU} &  \(7.8 \pm 1.6\)  & \(7.2 \pm 1.7\)  & \(6.0 \pm 1.3\)  & \(5.9 \pm 1.2\)  & \(7.5 \pm 1.4\)  & \(8.6 \pm 0.4\) \\

\multicolumn{1}{>{\columncolor{white}}l}{} & \multicolumn{2}{l}{\cellcolor{lightgray}Transformer} & \cellcolor{lightgray}\(14.6 \pm 1.1\) & \cellcolor{lightgray}\(8.9 \pm 1.0\) & \cellcolor{lightgray}\(20.8 \pm 3.1\) & \cellcolor{lightgray}\(8.2 \pm 1.4\)   & \cellcolor{lightgray}\(18.4 \pm 2.2\) & \cellcolor{lightgray}\(9.6 \pm 1.2\)  \\
 
 & \multicolumn{2}{l}{FA-Transformer} & \(6.9 \pm 1.2\)  & \(8.7 \pm 0.3\)  & \(9.9 \pm 4.1\) & \(8.1 \pm 0.5\)  & \(9.2 \pm 4.0\) & \(8.1 \pm 2.4\) \\

 \multicolumn{1}{>{\columncolor{white}}l}{}&  \cellcolor{lightgray}TFT & \cellcolor{lightgray} & \cellcolor{lightgray}\(20.8 \pm 4.1\) & \cellcolor{lightgray}\(25.4 \pm 2.9\) & \cellcolor{lightgray}\(22.7 \pm 4.5\) & \cellcolor{lightgray}\(22.3 \pm 3.7\) & \cellcolor{lightgray}\(7.8 \pm 0.8\) & \cellcolor{lightgray}\(8.6 \pm 0.5\) \\
 
\multicolumn{1}{>{\columncolor{white}}l}{\multirow{-11}{*}{\rotatebox[origin=c]{90}{\textbf{Non-Hierarchical}}}} & N-BEATS    &               & \(8.1 \pm 1.2\)  & \underline{\(6.2 \pm 0.4\)}  & \(8.1 \pm 2.0\)  & \(5.5 \pm 0.4\)  & \(10.9 \pm 1.8\) & \(8.7 \pm 0.6\)\\
 
 \arrayrulecolor{black}\midrule
 \multirow{10}{*}{\rotatebox[origin=c]{90}{\textbf{Hierarchical}}} 
 & \multicolumn{1}{>{\columncolor{lightgray}}l}{}& \cellcolor{lightgray}Na\"ive        & \cellcolor{lightgray}\color{blue}\textbf{---}  & \cellcolor{lightgray}8.0 & \cellcolor{lightgray}\color{blue}\textbf{---}   & \cellcolor{lightgray}7.6  & \cellcolor{lightgray}\color{blue}\textbf{---}   & \cellcolor{lightgray}\underline{7.6}  \\
  
 & \multicolumn{1}{>{\columncolor{lightgray}}l}{} & ARIMA            & \color{blue}\textbf{---}  & 11.0 & \color{blue}\textbf{---}  & 10.5  & \color{blue}\textbf{---}  & 10.5 \\
 
 & \multicolumn{1}{>{\columncolor{lightgray}}l}{\multirow{-3}{*}{Top-Down}}  & \cellcolor{lightgray}ETS              & \cellcolor{lightgray}\color{blue}\textbf{---}  & \cellcolor{lightgray}12.0 & \cellcolor{lightgray}\color{blue}\textbf{---}  & \cellcolor{lightgray}11.5  & \cellcolor{lightgray}\color{blue}\textbf{---}  & \cellcolor{lightgray}11.5  \\

& \multirow{3}{*}{Middle-Out}   & Na\"ive  & \color{blue}\textbf{---}  & 8.0 & \color{blue}\textbf{---}  & 7.6  & \color{blue}\textbf{---}  & \underline{7.6}\\
  
 &  & \cellcolor{lightgray}ARIMA          & \cellcolor{lightgray}\color{blue}\textbf{---}  & \cellcolor{lightgray}10.5 & \cellcolor{lightgray}\color{blue}\textbf{---}  & \cellcolor{lightgray}9.8  & \cellcolor{lightgray}\color{blue}\textbf{---}  & \cellcolor{lightgray}9.8   \\
 
 & & ETS            & \color{blue}\textbf{---} & 11.0 & \color{blue}\textbf{---}  & 10.4  & \color{blue}\textbf{---}  & 10.4 \\
 
 & \multicolumn{1}{>{\columncolor{lightgray}}l}{} & \cellcolor{lightgray}Na\"ive       & \cellcolor{lightgray}\color{blue}\textbf{---}  & \cellcolor{lightgray}8.0 & \cellcolor{lightgray}\color{blue}\textbf{---}  & \cellcolor{lightgray}7.6  & \cellcolor{lightgray}\color{blue}\textbf{---}  & \cellcolor{lightgray}\underline{7.6}  \\
 
  & \multicolumn{1}{>{\columncolor{lightgray}}l}{} &ARIMA           & \color{blue}\textbf{---}  & 9.7 & \color{blue}\textbf{---}  &  9.4  & \color{blue}\textbf{---}  & 9.4 \\
 
 & \multicolumn{1}{>{\columncolor{lightgray}}l}{\multirow{-3}{*}{Bottom-Up}}  &\cellcolor{lightgray}ETS             & \cellcolor{lightgray}\color{blue}\textbf{---}  & \cellcolor{lightgray}10.2 & \cellcolor{lightgray}\color{blue}\textbf{---}  &  \cellcolor{lightgray}9.2  & \cellcolor{lightgray}\color{blue}\textbf{---}  & \cellcolor{lightgray}9.2  \\
 
\cline{2-9}
& \multicolumn{2}{l}{\textbf{HierSTT (ours)}}& \color{Green}\textbf{\(2.0 \pm 0.4\)} & \color{Green}\textbf{\(5.6 \pm 0.2\)}  & \color{Green}\textbf{\(3.1 \pm 0.7 \)} & \color{Green}\textbf{\(4.9 \pm 0.2\)}  & \underline{\(3.3 \pm 0.7\)}  & \color{Green}\textbf{\(4.3 \pm 0.2\)} \\

\bottomrule
\end{tabular}
}
\end{table*}

Table~\ref{tab:hage_results} reports HAgE across the three aggregation transitions. Prediction-side HAgE measures inconsistencies between forecasts at different hierarchy levels. Na\"ive forecasts are implicitly coherent because repeating historical observations preserves the aggregation structure, while reconciliation baselines achieve zero prediction-side HAgE through explicit post-hoc aggregation constraints. These values are therefore marked with `\textcolor{blue}{\textbf{---}}', since coherence is algebraically enforced rather than learned. 

Among statistical and deep learning models, the standard Transformer exhibits the worst coherence (14.6\%, 20.8\%, 18.4\%), whereas ARIMA achieves the lowest prediction-side HAgE (3.7\%, 3.9\%, 2.1\%), reflecting its smoother forecasts. In contrast, HierSTT learns hierarchical consistency directly through \(\mathcal{L}_{coh}\), achieving prediction-side HAgE of 2.0\%, 3.1\%, and 3.3\%, corresponding to the best results for hospital-to-regional and hospital-to-national aggregation, and second best for regional-to-national.

Ground-truth HAgE measures the discrepancy between aggregated lower-level predictions and observed higher-level values. HierSTT also achieves the lowest errors across all transitions (5.6\%, 4.9\%, 4.3\%), indicating that its lower-level forecasts remain both coherent and accurate when aggregated, unlike reconciliation baselines that enforce coherence but remain limited by the quality of their base forecasts.

\subsection{Qualitative Forecast Results}
\begin{figure}[t]
    \centering
    \includegraphics[width=0.8\linewidth]{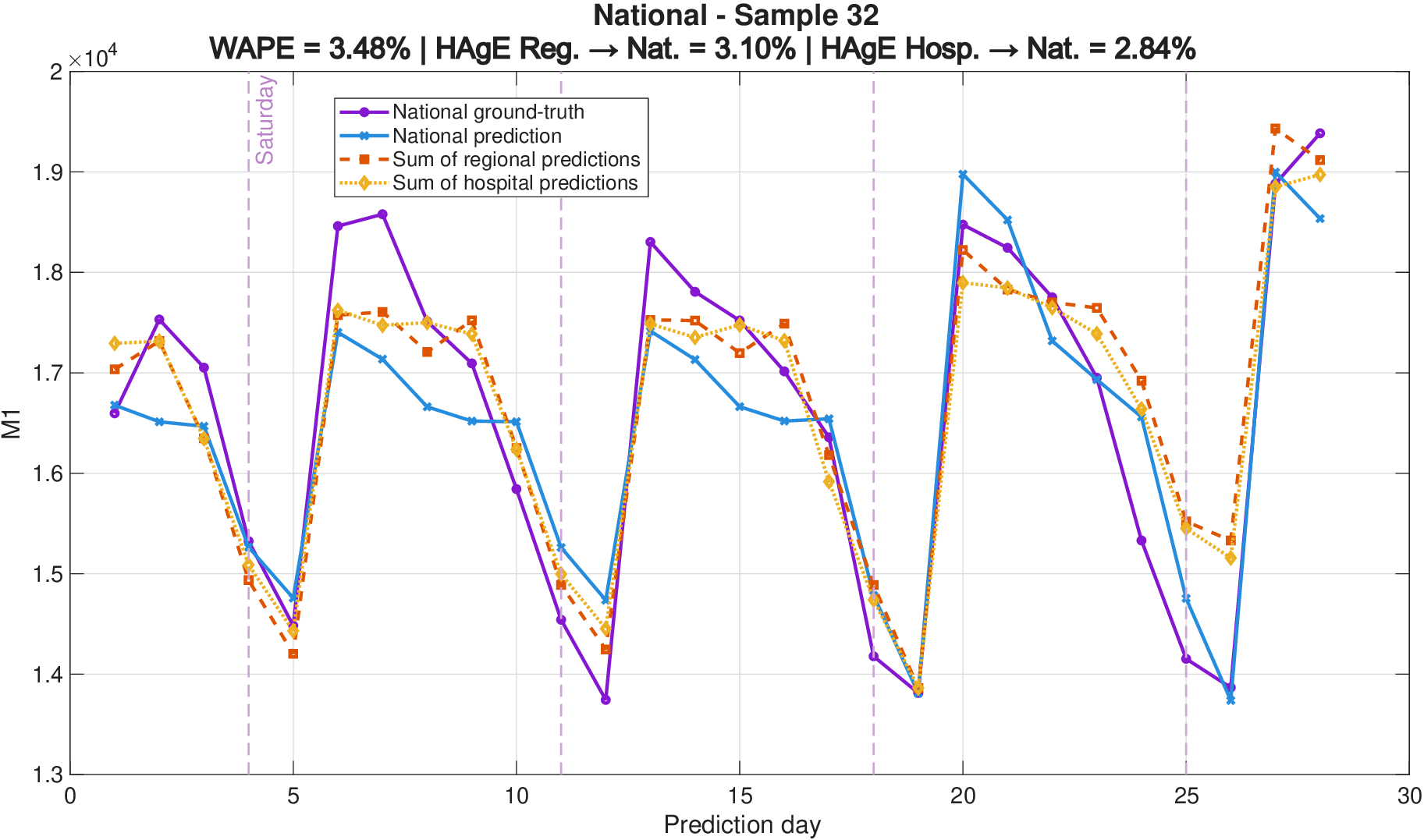}
    \caption{Example of the national-level forecast for the test sample with the lowest national WAPE. The figure compares the ground truth, the direct national prediction, the sum of regional predictions, and the sum of hospital predictions over the 28-day forecasting horizon. Vertical lines indicate Saturdays.}
    \label{fig:best_nat}
\end{figure}

\begin{figure}[t]
    \centering
    \includegraphics[width=\linewidth]{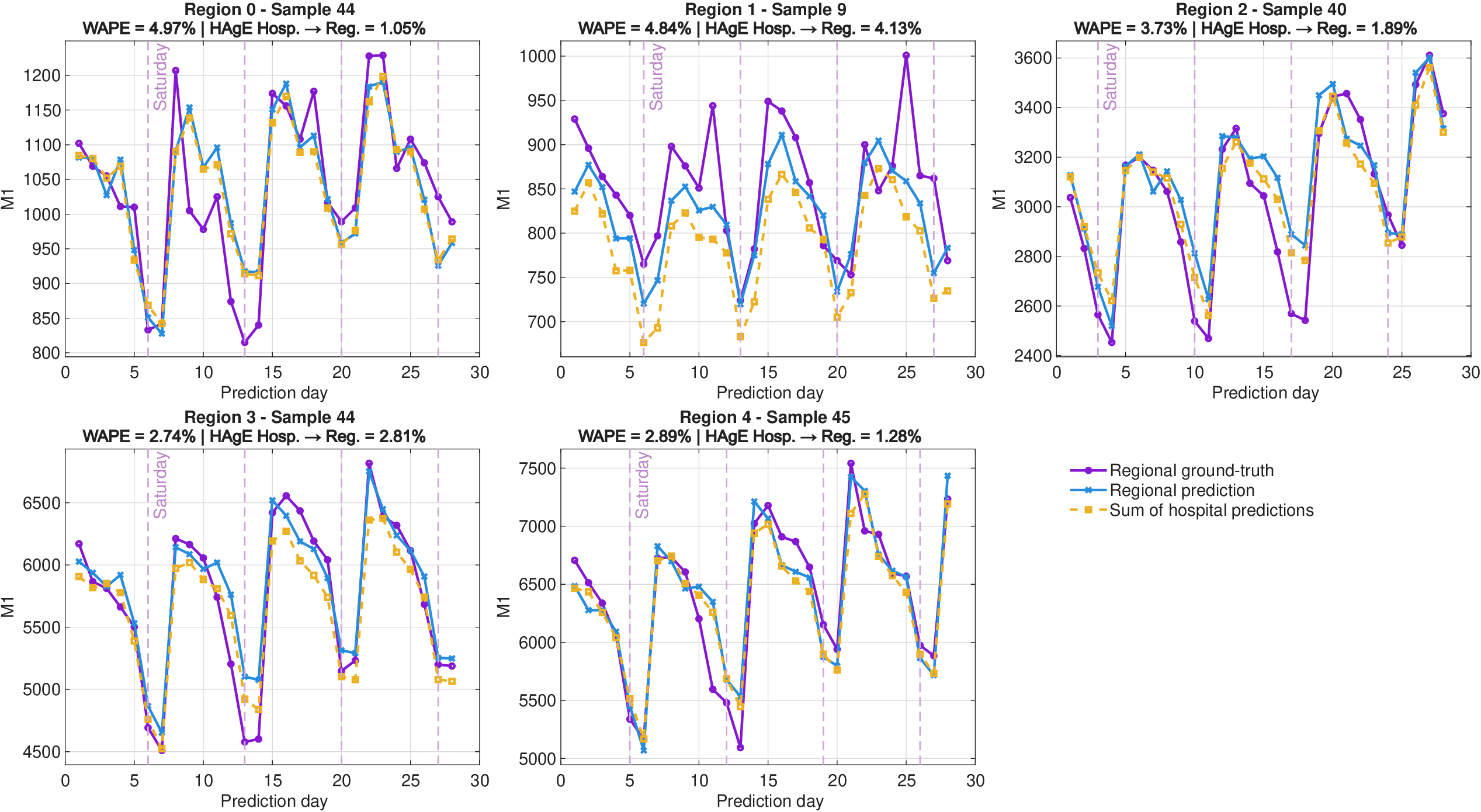}
    \caption{Regional-level forecasts for the test sample with the lowest WAPE independently selected for each region. Each subplot shows the regional ground truth, the direct regional prediction, and the sum of the corresponding hospital-level predictions. Vertical lines indicate Saturdays.}
    \label{fig:best_reg}
\end{figure}

\begin{figure}[t]
    \centering
    \begin{subfigure}{0.49\textwidth}
        \centering
        \includegraphics[width=\textwidth]{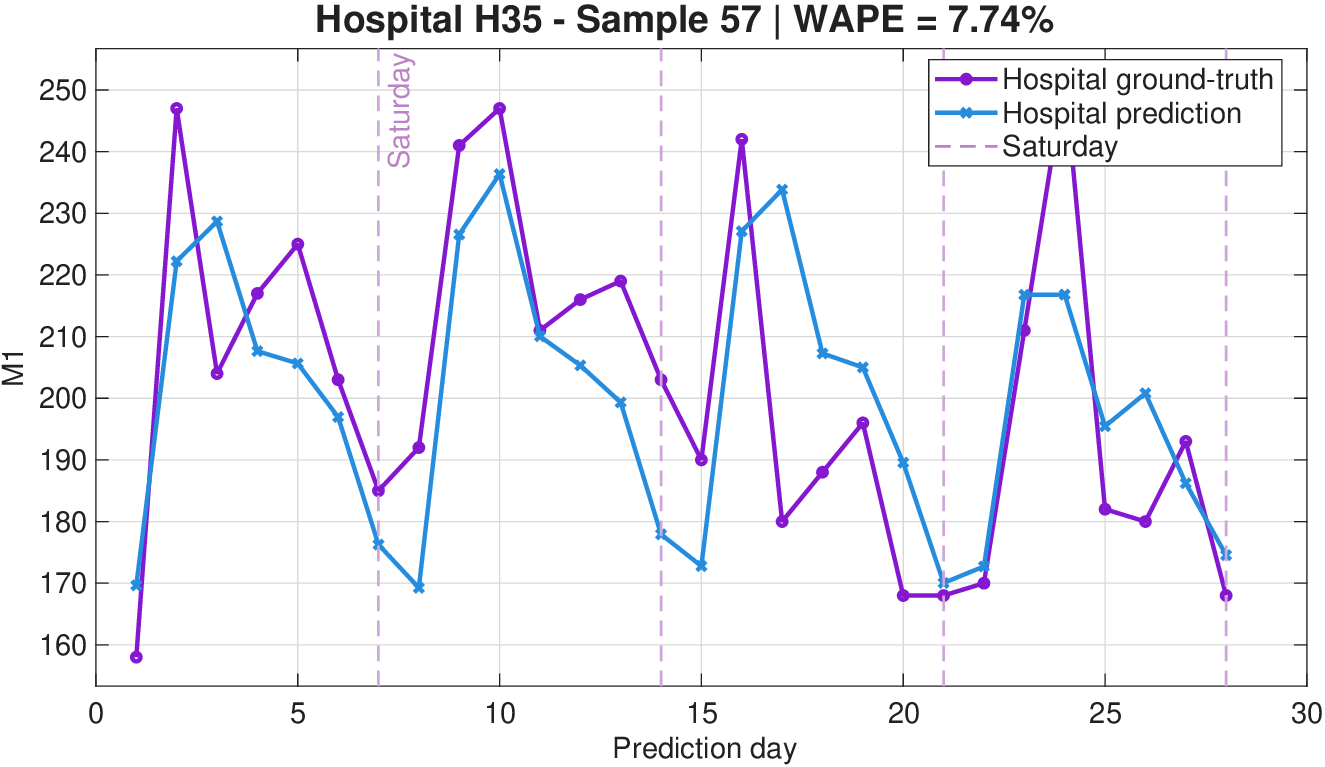}
        \caption{}
        \label{fig:best_hosp_reg_0}
    \end{subfigure}
    \hfill
    \begin{subfigure}{0.49\textwidth}
        \centering
        \includegraphics[width=\textwidth]{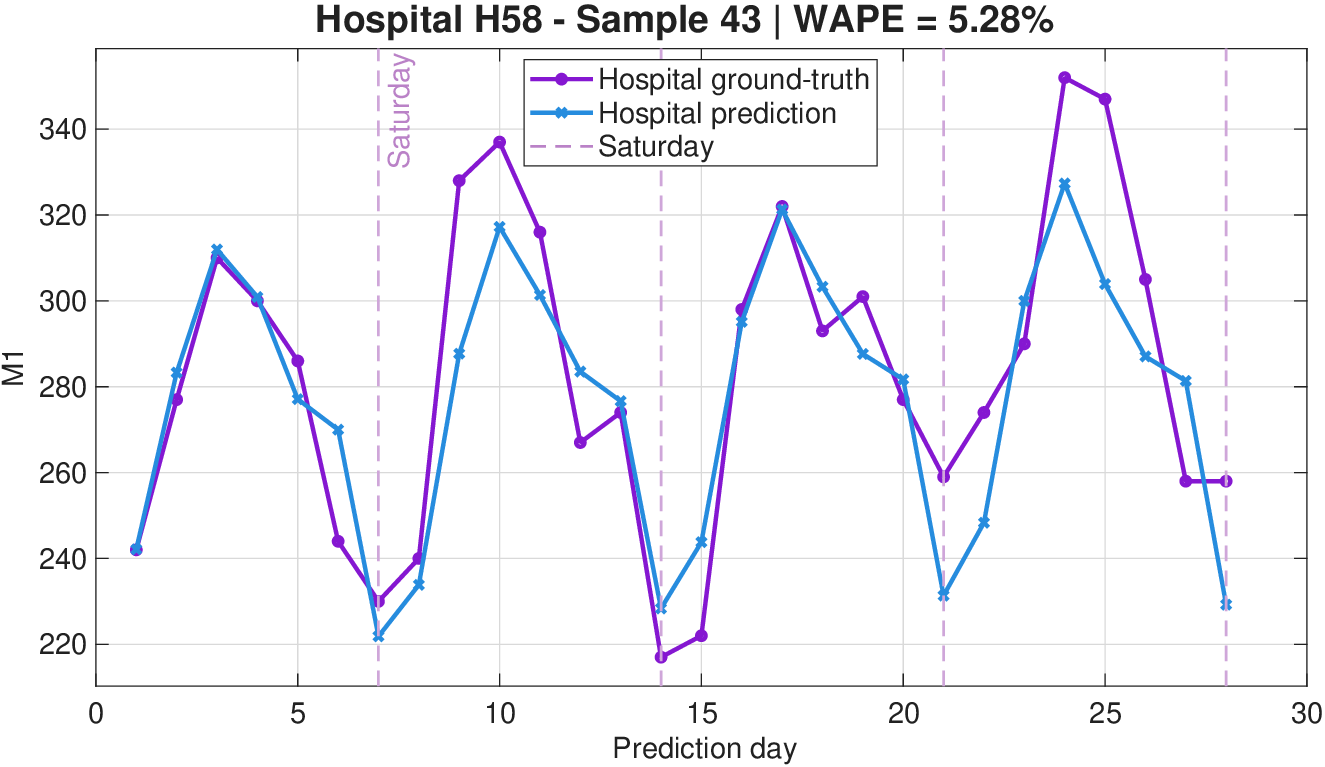}
        \caption{}
        \label{fig:best_hosp_reg_1}
    \end{subfigure}
    \hfill
    \begin{subfigure}{0.49\textwidth}
        \centering
        \includegraphics[width=\textwidth]{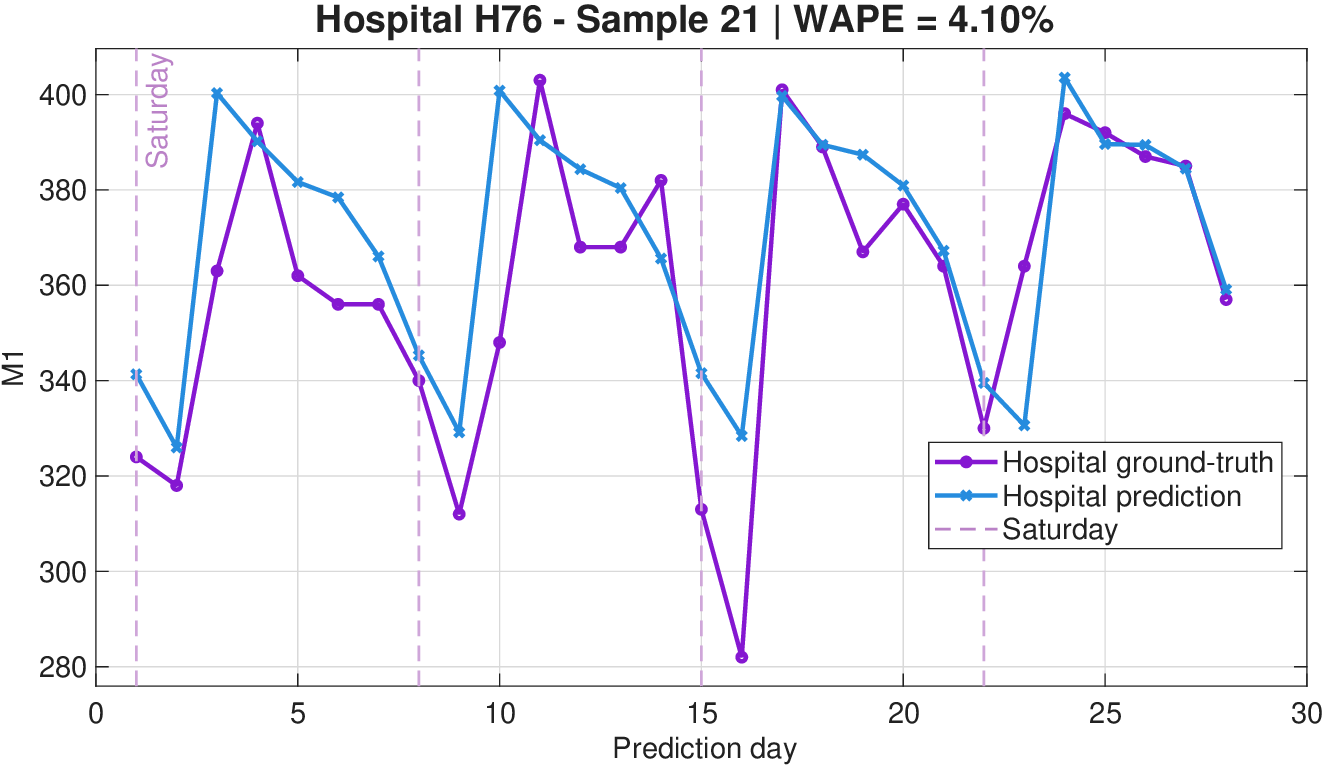}
        \caption{}
        \label{fig:best_hosp_reg_2}
    \end{subfigure}
    \hfill
    \begin{subfigure}{0.49\textwidth}
        \centering
        \includegraphics[width=\textwidth]{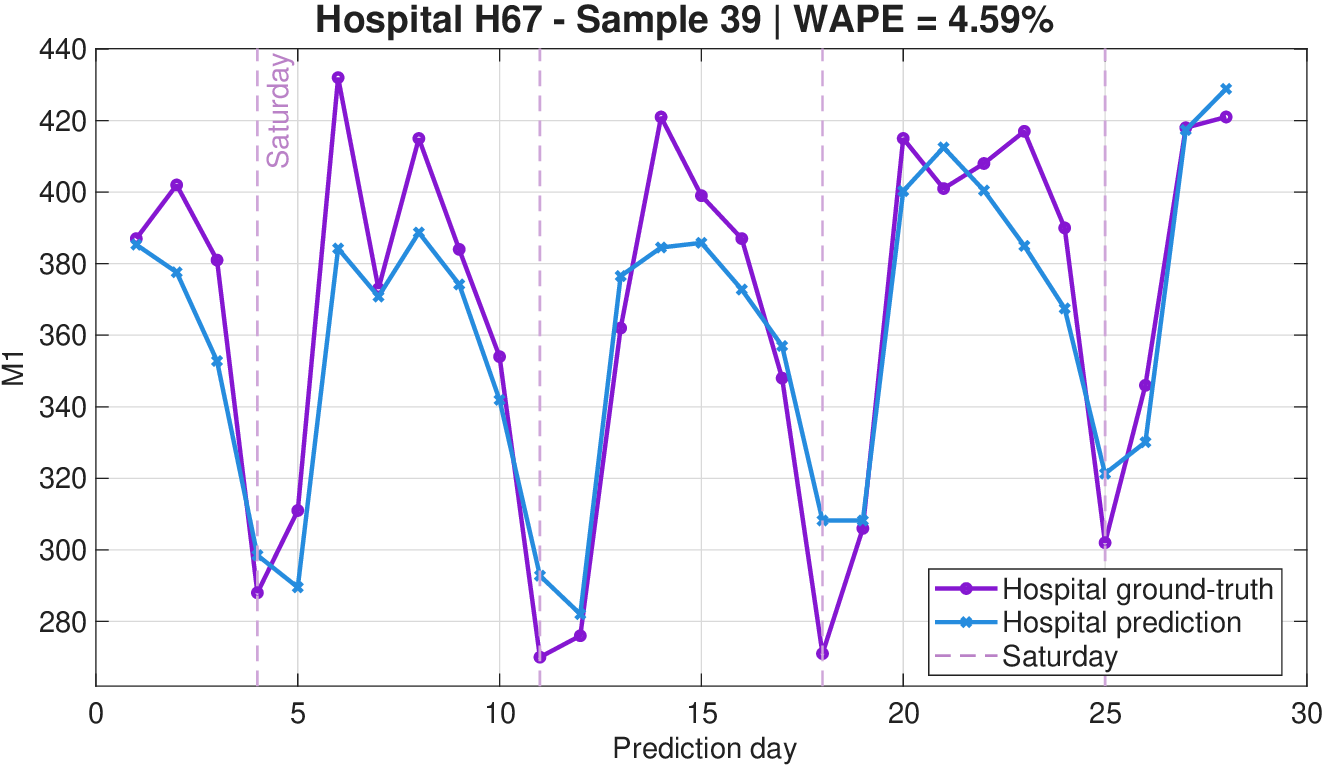}
        \caption{}
        \label{fig:best_hosp_reg_3}
    \end{subfigure}
    \hfill
    \begin{subfigure}{0.49\textwidth}
        \centering
        \includegraphics[width=\textwidth]{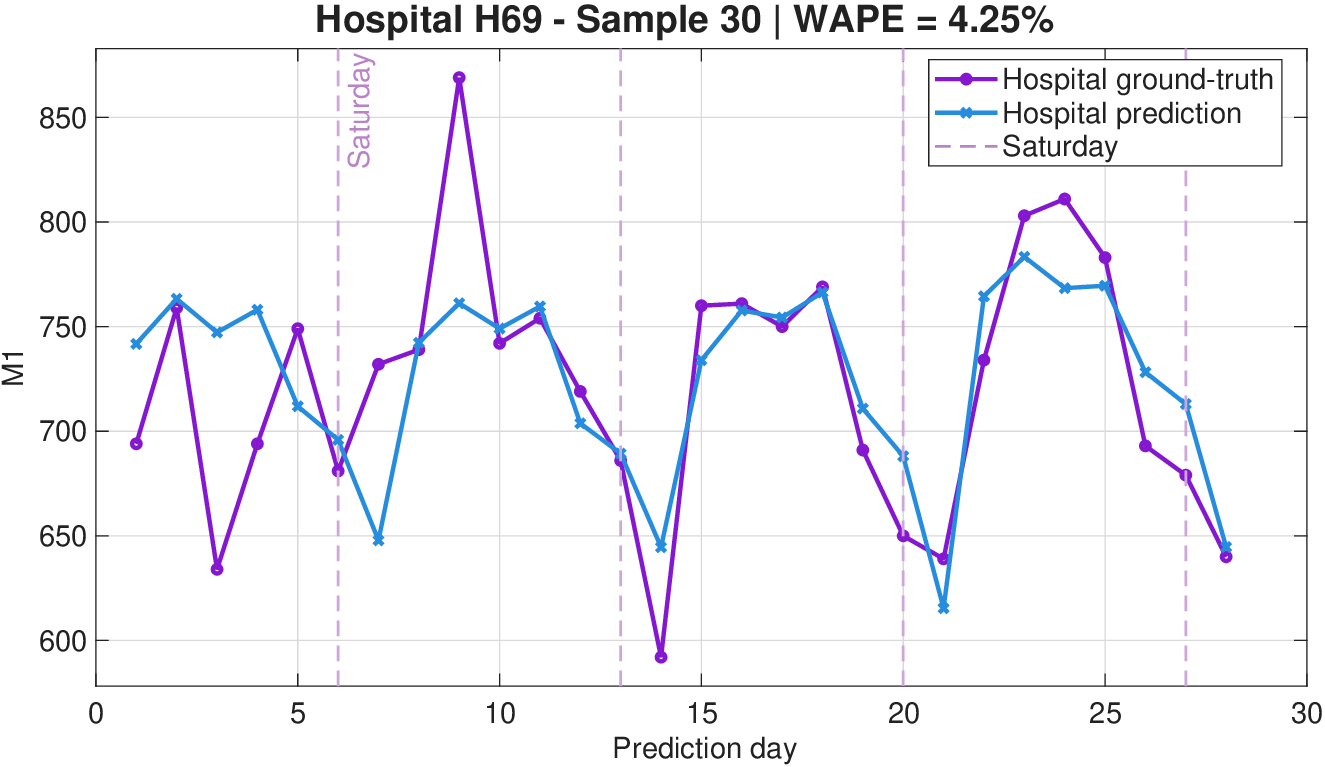}
        \caption{}
        \label{fig:best_hosp_reg_4}
    \end{subfigure}
        \hfill
    \begin{subfigure}{0.49\textwidth}
        \centering
        \includegraphics[width=\textwidth]{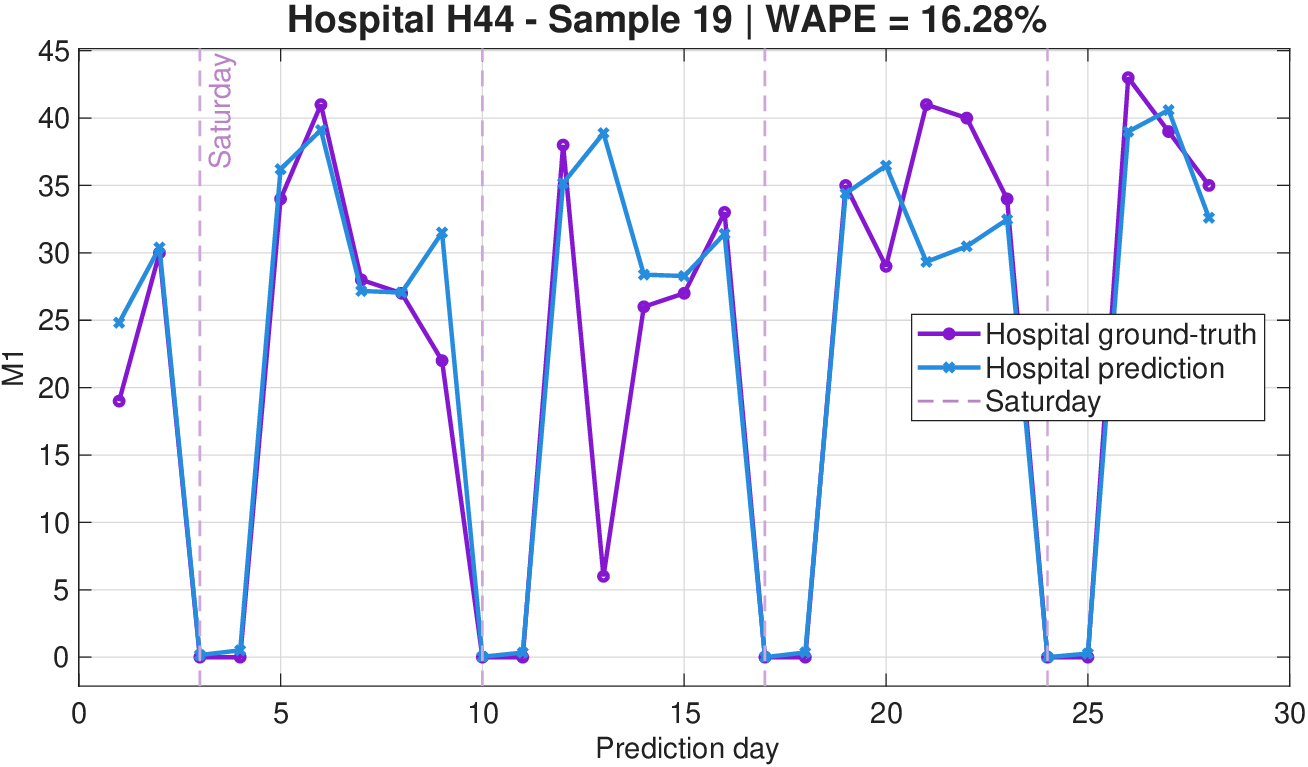}
        \caption{}
        \label{fig:hosp_zeros}
    \end{subfigure}
    \caption{Hospital-level forecasts for representative hospitals selected from each region. For each region, the hospital example with the lowest WAPE was selected ((a) to (e)), and an additional example is shown for Hospital H44 (f), which includes days in which the hospital is closed. Vertical lines indicate Saturdays.}
    \label{fig:best_hospitals}
\end{figure}

To complement the quantitative evaluation, we provide qualitative forecast examples generated by HierSTT in Fig.~\ref{fig:best_nat},~\ref{fig:best_reg}, and~\ref{fig:best_hospitals}. Examples were selected based on the lowest WAPE at each hierarchy level, allowing inspection of both forecasting accuracy and consistency between direct predictions and hierarchical aggregations. The added vertical Saturday markers further highlight the weekly structure present across hierarchy levels, supporting the inclusion of calendar-related covariates.

At the national and regional levels (Fig.~\ref{fig:best_nat} and~\ref{fig:best_reg}), the model captures the weekly demand structure and the main rises and drops across the 28-day horizon. The aggregated lower-level predictions closely follow the direct forecasts, indicating strong hierarchical coherence throughout the forecasting window. In particular, the national example  (Fig.~\ref{fig:best_hospitals}) achieves a WAPE of 3.48\%, while remaining highly consistent with both aggregated regional and hospital predictions.

Across the five regional examples (Fig.~\ref{fig:best_reg}), the aggregated hospital predictions closely follow the regional forecasts. Regional differences may partly reflect the hierarchical composition of each region. Regions with more hospitals aggregate a larger number of local demand signals, smoothing hospital-specific variability and facilitating coherent regional prediction. In contrast, regions with fewer hospitals, such as Region 0 and 1, are more sensitive to individual hospital fluctuations.

At the hospital level (Fig.~\ref{fig:best_hospitals}), the model captures local temporal dynamics despite the higher variability of individual hospital demand. Some sharp peaks are underestimated, particularly in lower-volume hospitals (Fig.~\ref{fig:best_hosp_reg_0} and Fig.~\ref{fig:best_hosp_reg_1}), but the main temporal patterns are preserved. We additionally include Hospital H44 (Fig.~\ref{fig:hosp_zeros}) as an operational example containing hospital closure days. The model predicts near-zero demand during these periods, illustrating the benefit of including the hospital open/closed status as a future-known decoder variable.

\end{document}